\pdfoutput=1

\documentclass{article}

\usepackage[nonatbib,final]{neurips_2019}

\usepackage[utf8]{inputenc} 
\usepackage[T1]{fontenc}    
\usepackage{hyperref}       
\usepackage{url}            
\usepackage{booktabs}       
\usepackage{amsfonts}       
\usepackage{nicefrac}       
\usepackage{microtype}      

\usepackage{epsfig}
\usepackage{algorithm}
\usepackage{algorithmic}
\usepackage{xcolor}
\usepackage{amsmath,amssymb}
\usepackage{booktabs,subcaption,dcolumn}

\newif\ifsubmit
\submitfalse

\definecolor{medGray}{RGB}{230,230,230}

\makeatletter
\newcommand{\figcaption}[1]{\def\@captype{figure}\caption{#1}}
\newcommand{\tblcaption}[1]{\def\@captype{table}\caption{#1}}
\makeatother

\ifsubmit
\newcommand\todo[1]{}
\newcommand\kwrite[1]{}
\newcommand\mwrite[1]{}
\newcommand\mnote[1]{}
\newcommand\ynote[1]{}
\newcommand\knote[1]{}
\else
\newcommand\todo[1]{\textcolor{red}{TODO: #1}}
\newcommand\mnote[1]{\textcolor{blue}{(MIYATO: #1)}}
\newcommand\mwrite[1]{\textcolor{green}{#1}}
\newcommand\kwrite[1]{\textcolor{brown}{#1}}
\newcommand\ynote[1]{\textcolor{blue}{(Yoshida: #1)}}
\newcommand\knote[1]{\textcolor{blue}{(Kataoka: #1)}}
\fi

\newcommand{\bbR}{\mathbb{R}}

\graphicspath{{figures/}}

\newtheorem{proposition}{Proposition}
\newtheorem{theorem}{Theorem}
\newtheorem{corollary}{Corollary}

\title{Self-supervised GAN: Analysis and Improvement with Multi-class Minimax Game}

\author{%
  Ngoc-Trung Tran, Viet-Hung Tran, Ngoc-Bao Nguyen, Linxiao Yang, Ngai-Man Cheung\\
  Singapore University of Technology and Design (SUTD)\\
  \\
  Corresponding author: Ngai-Man Cheung \texttt{<ngaiman\_cheung@sutd.edu.sg>}\\
}

\begin{document}

\maketitle

\begin{abstract}

Self-supervised (SS) learning is a powerful approach for representation learning using unlabeled data. Recently, it has been applied to Generative Adversarial Networks (GAN) training. Specifically, SS tasks were proposed to address the catastrophic forgetting issue in the GAN discriminator. In this work, we perform an in-depth analysis to understand how SS tasks interact with learning of generator. From the analysis, we identify issues of SS tasks which allow a severely mode-collapsed generator to excel the SS tasks. To address the issues, we propose new SS tasks based on a multi-class minimax game. The competition between our proposed SS tasks in the game encourages the generator to learn the data distribution and generate diverse samples. We provide both theoretical and empirical analysis to support that our proposed SS tasks have better convergence property. We conduct experiments to incorporate our proposed SS tasks into two different GAN baseline models. Our approach establishes state-of-the-art FID scores on CIFAR-10, CIFAR-100, STL-10, CelebA, Imagenet $32\times32$ and Stacked-MNIST datasets, outperforming existing works by considerable margins in some cases. Our unconditional GAN model approaches performance of conditional GAN {\em without} using labeled data.  Our code:  \url{https://github.com/tntrung/msgan}



\end{abstract}

\section{Introduction}

{\bf Generative Adversarial Networks (GAN).} GAN~\cite{goodfellow-nisp-2014} have become one of the most important methods to learn generative models. GAN has shown remarkable results in various tasks, such as: image generation \cite{karras-iclr-2018,brock-iclr-2018,karras-cvpr-2019}, image transformation \cite{isola-cvpr-2017, zhu-cvpr-2017}, super-resolution \cite{ledig-cvpr-2017}, text to image \cite{reed-arxiv-2016,zhang2-cvpr-2017}, anomaly detection \cite{schlegl-ipmi-2017,lim-icdm-2018}. The idea behind GAN is the mini-max game. It uses a binary classifier, so-called the discriminator, to distinguish the data (real) versus generated (fake) samples. 
The generator of GAN is trained to confuse the discriminator to classify the generated samples as the real ones. 
By having the generator and discriminator competing with each other in this adversarial process, they are able to improve themselves. The end goal is to have the generator capturing  
the data distribution. Although considerable improvement has been made for GAN under the conditional settings \cite{odena-icml-2017,zhang-arxiv-2018,brock-iclr-2018}, i.e., using ground-truth labels to support the learning, it is still very challenging with unconditional setup. Fundamentally, using only a single signal (real/fake) to guide the generator to learn the high-dimensional, complex data distribution is very challenging \cite{goodfellow-nips-2016,arjovsky-arxiv-2017a,che-arxiv-2016,chen-arxiv-2016,metz-arxiv-2016,salimans-nisp-2016}.

{\bf Self-supervised Learning.}
Self-supervised learning is an active research area \cite{doersch-cvpr-2015,pathak-cvpr-2016,zhang-eccv-2016,zhang1-cvpr-2017,noroozi-iccv-2017,gidaris-iclr-2018}. Self-supervised learning is a paradigm of unsupervised learning.
Self-supervised methods
encourage the classifier to learn better feature representation with {\em pseudo-labels}. In particular, these methods propose to learn image feature by training the model to recognize some geometric transformation that is applied to the image which the model receives as the input.
A simple-yet-powerful method proposed in \cite{gidaris-iclr-2018}
is to use image rotations by 0, 90, 180, 270 degrees as the geometric transformation.
The model is trained with the 4-way classification task of  recognizing one of the four rotations. This task is referred as 
the {\em self-supervised task}. This simple method is able to close the gap between supervised and unsupervised image classification \cite{gidaris-iclr-2018}. 

\begin{figure}
  \centering
  \includegraphics[scale=0.7]{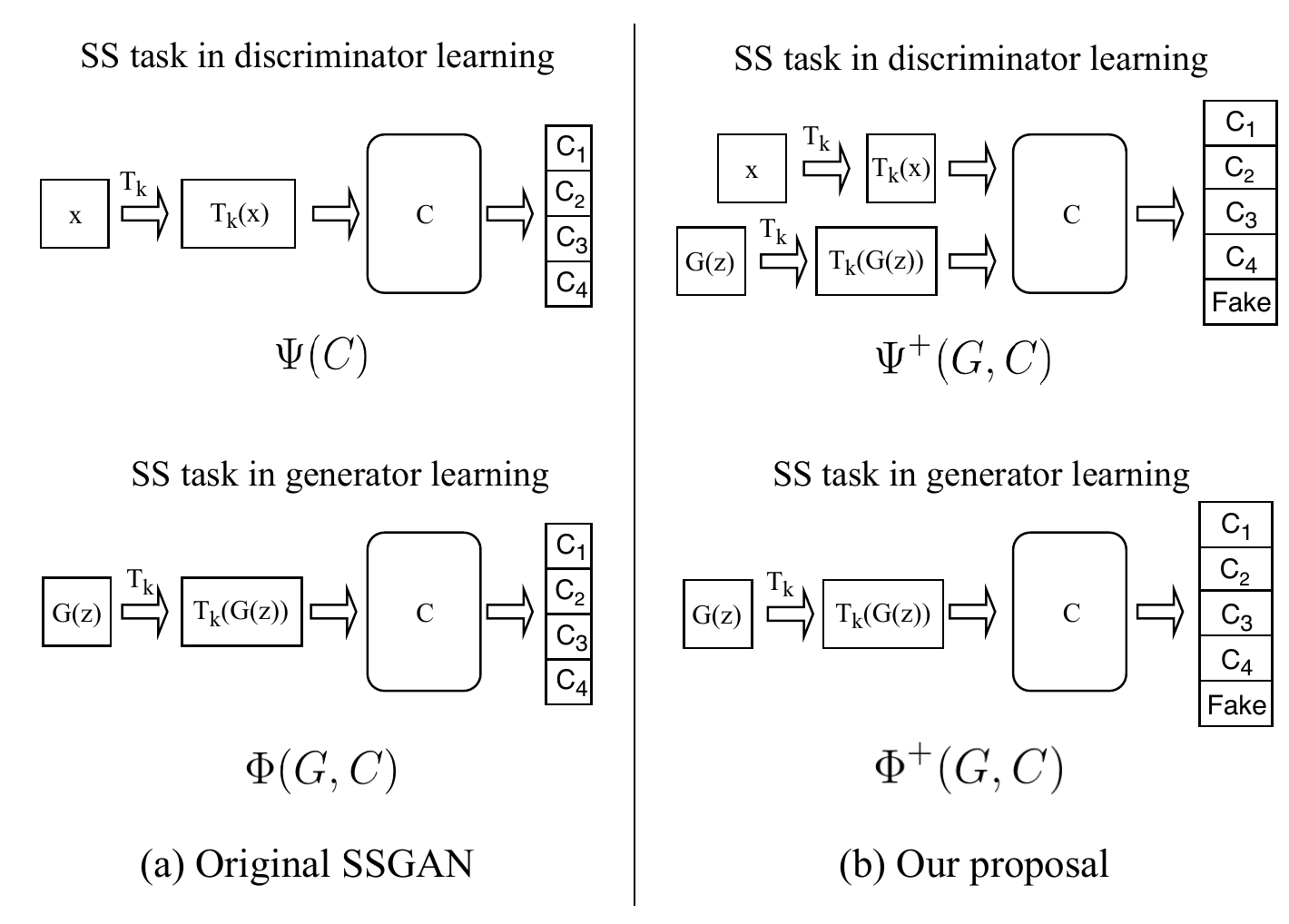}
  \caption{The model of (a) SSGAN \cite{chen-arxiv-2018} and (b) our approach. Here, $\Psi(C)$ and $\Phi(G,C)$ are the self-supervised value functions in training discriminator and generator, respectively, as proposed in \cite{chen-arxiv-2018}. $\Psi^+(G,C)$ and $\Phi^+(G,C)$ are the self-supervised value functions proposed in this work.}
  \label{proposed_model}
  \vspace{-0.4cm}
\end{figure}

{\bf Self-supervised Learning for GAN.}
Recently, self-supervised learning has been applied to GAN training
\cite{chen-arxiv-2018,tran-arxiv-2019}.  These works propose 
auxiliary 
self-supervised classification tasks to assist the main GAN task (Figure~\ref{proposed_model}).
In particular, their objective functions for learning discriminator $D$ and 
generator $G$  are 
multi-task loss as shown in (\ref{gan_dis_obj_ss1}) and (\ref{gan_gen_obj_ss1}) respectively:
\begin{equation}
\begin{split}
\max_{D,C}\mathcal{V_D}(D,C,G) &= \mathcal{V}(D,G) + \lambda_d {\Psi(G,C)} \\
\end{split}
\label{gan_dis_obj_ss1}
\end{equation}
\vspace{-0.5cm}
\begin{equation}
\begin{split}
\min_G\mathcal{V_G}(D,C,G) &= \mathcal{V}(D,G) - \lambda_g {\Phi(G,C)} \\
\end{split}
\label{gan_gen_obj_ss1}
\end{equation}
\vspace{-0.5cm}
\begin{equation}
\begin{split}
\mathcal{V}(D,G) &=  \mathbb{E}_{\mathbf{x} \sim {P_d}}\log \Big(D(\mathbf{x})\Big)   + \mathbb{E}_{\mathbf{x} \sim {P_g}}\log \Big(1-D(\mathbf{x})\Big)
\end{split}
\label{gan_original}
\end{equation}
Here, $\mathcal{V}(D,G)$ in (\ref{gan_original}) is the {\em GAN task}, which is the original value function proposed in Goodfellow et al.~\cite{goodfellow-nisp-2014}.
$P_d$ is true data distribution, $P_g$ is the distribution induced by the generator mapping.
$\Psi(G,C)$ and $\Phi(G,C)$ are the {\em self-supervised (SS) tasks} for discriminator and generator learning, respectively (details to be discussed). $C$ is the classifier for the self-supervised task, e.g. rotation classifier 
as discussed \cite{gidaris-iclr-2018}.
Based on this framework, Chen et al.\cite{chen-arxiv-2018} apply self-supervised task to help discriminator counter catastrophic forgetting. Empirically, they have shown that self-supervised task enables discriminator to learn more stable and improved representation. Tran et al.~\cite{tran-arxiv-2019} propose to improve self-supervised learning with adversarial training. 

Despite the encouraging empirical results, in-depth analysis of the interaction between SS tasks ($\Psi(.)$ and $\Phi(.)$) and GAN task ($\mathcal{V}(D,G)$) has not been done before.
On one hand, the application of SS task for {\em discriminator learning} is reasonable: the goal of discriminator is to classify real/fake image; an additional SS classification task $\Psi(G,C)$ could assist feature learning and enhance the GAN task. On the other hand, the motivation and design of SS task for {\em generator learning} is rather subtle: the goal of generator learning is to capture the data distribution in $G$, and it is unclear exactly how an additional SS classification task $\Phi(G,C)$ could help.



{\bf In this work,} we conduct in-depth empirical and theoretical analysis to understand the interaction between self-supervised tasks ($\Psi(.)$ and $\Phi(.)$) and learning of generator $G$.
Interestingly, from our analysis,
we reveal issues of existing works. 
Specifically, 
the SS tasks of existing works have ``loophole'' that, during generator learning, $G$ could exploit to maximize $\Phi(G,C)$ without truly learning the data distribution. We show that analytically and empirically that a severely mode-collapsed generator can excel $\Phi(G,C)$. 
To address this issue, we
propose new SS tasks based on a multi-class minimax game.
Our proposed new SS tasks of discriminator and generator 
compete with each other to reach the equilibrium point. Through this competition, our proposed SS tasks are able to support the GAN task better. Specifically, our analysis shows that {\em our proposed SS tasks enhance matching between $P_d$ and $P_g$ by leveraging the transformed samples used in the SS classification} (rotated images when 
\cite{gidaris-iclr-2018} is applied).
In addition, our design couples GAN task and SS task. 
To validate our design, 
we provide theoretical analysis on the convergence property of our proposed SS tasks. Training a GAN with our proposed self-supervised tasks based on multi-class minimax game significantly improves baseline models. Overall, our system  establishes state-of-the-art Fr{\'e}chet Inception Distance (FID) scores.
In summary, our contributions are:
\begin{itemize}
	\item We conduct in-depth empirical and theoretical analysis to understand the issues of self-supervised tasks in existing works.
	\item Based on the analysis, we propose new self-supervised tasks based on a multi-class minimax game.
	\item We conduct extensive experiments to validate our proposed self-supervised tasks.
\end{itemize}

\section{Related works}

While training GAN with conditional signals (e.g., ground-truth labels of classes) has made good progress \cite{odena-icml-2017,zhang-arxiv-2018,brock-iclr-2018}, training GAN in the unconditional setting is still very challenging. In the original GAN \cite{goodfellow-nisp-2014}, the single signal (real or fake) of samples is provided to train discriminator and the generator. With these signals, the generator or discriminator may fall into ill-pose settings, and they may get stuck at bad local minimums though still satisfying the signal constraints. To overcome the problems, many regularizations have been proposed. One of the most popular approaches is to enforce (towards) Lipschitz condition of the discriminator. These methods include weight-clipping \cite{arjovsky-arxiv-2017a}, gradient penalty constraints \cite{gulrajani-arxiv-2017,roth-nips-2017,kodali-arxiv-2017,petzka-arxiv-2017,liu-arxiv-2018} and spectral norm \cite{miyato-iclr-2018}. Constraining the discriminator mitigates gradients vanishing and avoids sharp decision boundary between the real and fake classes.


Using Lipschitz constraints improve the stability of GAN. However, the challenging optimization problem still remains when using a single supervisory signal, similar to the  original GAN \cite{goodfellow-nisp-2014}. In particular, the learning of discriminator is highly dependent on generated samples. If the generator collapses to some particular modes of data distribution, it is only able to create samples around these modes. 
There is no competition to train the discriminator around other modes. As a result, the gradients of these modes may vanish, and it is impossible for the generator to model well the entire data distribution. Using additional supervisory signals helps the optimization process. For example, using self-supervised learning in the form of auto-encoder has been proposed. AAE \cite{makhzani-arxiv-2015} guides the generator towards resembling realistic samples. 
However, an issue with using auto-encoder is that pixel-wise reconstruction with $\ell_2$-norm causes blurry artifacts. VAE/GAN \cite{larsen-arxiv-2015}, which combining  VAE \cite{kingma-arxiv-2013} and GAN, is an improved solution: while the discriminator of GAN enables the usage of feature-wise reconstruction to overcome the blur, the VAE constrains the generator to mitigate mode collapse. In ALI \cite{dumoulin-arxiv-2016} and BiGAN \cite{donahue-arxiv-2016}, they jointly train the data/latent samples in the GAN framework. InfoGAN \cite{chen-arxiv-2016} infers the disentangled latent representation by maximizing the mutual information. In \cite{tran-eccv-2018,tran-aaai-2018}, they  combine two different types of supervisory signals: real/fake signals and self-supervised signal in the form of auto-encoder.  In addition, Auto-encoder based methods, including \cite{larsen-arxiv-2015,tran-eccv-2018,tran-aaai-2018},  can be considered as an approach  to mitigate catastrophic forgetting because they  regularize the generator to resemble the real ones. It is similar to EWC \cite{kirkpatrick-2017-nas} or IS \cite{zenke-arxiv-2017} but the regularization is achieved via the output, not the parameter itself. Although using feature-wise distance in auto-encoder could reconstruct sharper images, it is  still challenging to produce very realistic detail of textures or shapes.

Several different types of supervisory signal have been proposed. Instead of using only one discriminator or generator, they propose ensemble models, such as multiple discriminators \cite{tu-nips-2017}, mixture of generators \cite{hoang-arxiv-2018,ghosh-cvpr-2018} or applying an attacker as a new player for GAN training \cite{liu-cvpr-2019}. Recently, training model with auxiliary self-supervised constraints \cite{chen-arxiv-2018,tran-arxiv-2019} via multi pseudo-classes \cite{gidaris-iclr-2018} helps improve stability of the  optimization process. This approach is appealing: it is simple to implement and does not require more parameters in the networks (except a small head for the classifier).
Recent work applies InfoMax principle to improve GAN \cite{kwotsin:2019}.  Variational Autoencoder is  another important approach to learn generative models \cite{kingma-arxiv-2013,Yang_2019_ICCV}.

\section{GAN with Auxiliary Self-Supervised tasks}

In \cite{chen-arxiv-2018}, self-supervised (SS) value function (also referred as ``self-supervised task'') was proposed for GAN \cite{goodfellow-nisp-2014} via image rotation prediction \cite{gidaris-iclr-2018}. 
In their work, they showed that the SS task was useful to mitigate catastrophic forgetting problem of GAN discriminator. The objectives of the discriminator and generator in \cite{chen-arxiv-2018} are shown in Eq. \ref{gan_dis_obj_ss} and \ref{gan_gen_obj_ss}. Essentially, the SS task of the discriminator (denoted by $\Psi(C)$) is to train the classifier $C$ that maximizes the performance of predicting 
the rotation applied to the {\em real} samples.
Given this classifier $C$, 
the SS task of the generator (denoted by $\Phi(G,C)$) is to train the generator $G$ to produce {\em fake} samples for maximizing classification performance. The discriminator and classifier are the same (shared parameters), except the last layer in order to implement  two different heads: the last fully-connected layer which returns a one-dimensional output (real or fake) for the discriminator,  and the other which returns a $K$-dimensional softmax of pseudo-classes for the classifier. $\lambda_d$ and $\lambda_g$ are constants.
\begin{equation}
\begin{split}
\max_{D,C}\mathcal{V}(D,C,G) &= \mathcal{V}(D,G) + \lambda_d \underbrace{\bigg(\mathbb{E}_{\mathbf{x} \sim {P_d^T}} \mathbb{E}_{T_k \sim \mathcal{T}}\log\Big(C_k(\mathbf{x})\Big)\bigg)}_{\Psi(C)}
\end{split}
\label{gan_dis_obj_ss}
\end{equation}
\begin{equation}
\begin{split}
\min_G\mathcal{V}(D,C,G) &= \mathcal{V}(D,G) - \lambda_g \underbrace{\bigg(\mathbb{E}_{\mathbf{x} \sim {P_g^T}} \mathbb{E}_{T_k \sim \mathcal{T}}\log\Big(C_k(\mathbf{x})\Big)\bigg)}_{\Phi(G,C)}
\end{split}
\label{gan_gen_obj_ss}
\end{equation}
Here, the GAN value function $\mathcal{V}(D,G)$ 
(also referred as ``GAN task'')
can be the original minimax GAN objective \cite{goodfellow-nisp-2014} or other improved versions.
$\mathcal{T}$ is the set of transformation, $T_k \in \mathcal{T}$ is the $k$-th transformation.
The rotation SS task proposed in \cite{gidaris-iclr-2018} is applied, and $T_1, T_2, T_3, T_4$ are the 0, 90, 180, 270 degree image rotation, respectively.
$P_d, P_g$ are the distributions of real and fake data samples, respectively.
$P_d^T, P_g^T$ are the mixture distribution of {\em rotated} real and fake data samples (by $T_k \in \mathcal{T}$), respectively.
Let $C_k(\mathbf{x})$ be the $k$-th softmax output of classifier $C$, and we have $\sum_{k=1}^{K}C_k(\mathbf{x}) = 1, \forall \mathbf{x}$. The models are shown in Fig. \ref{proposed_model}a. In \cite{chen-arxiv-2018}, empirical evidence of improvements has been provided.

Note that, the goal of $\Phi(G,C)$ is to encourage the generator to produce realistic images. It is because classifier $C$ is trained with real images and captures features that allow detection of rotation. However, the interaction of $\Phi(G,C)$ with the GAN task $\mathcal{V}(D,G)$ has not been adequately analyzed.

\section{Analysis on Auxiliary Self-supervised Tasks}
\label{analysis_on_auxiliary_ss}

\begin{figure}
  \centering
  \includegraphics[scale=0.43]{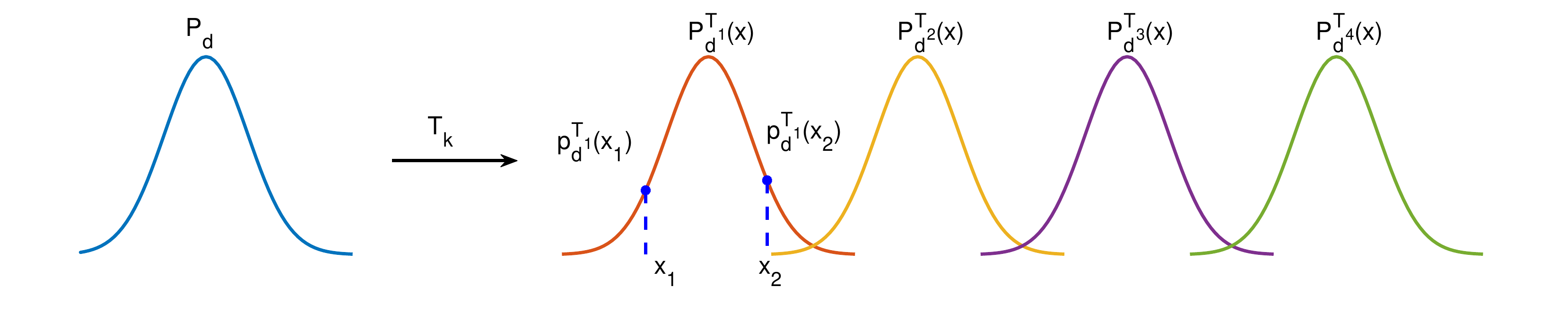}
  \caption{The probability distribution $p_d^{T_k}(\mathbf{x})$. Here, samples from $P_d$ are rotated by $T_k$. The distribution of rotated sample is $p^{T_k}(\mathbf{x})$.
  Some rotated samples resemble the original samples, e.g. those on the right of $\mathbf{x}_2$. On the other hand, for some image, there is no rotated image resembling it, e.g. $\mathbf{x}_1$ ($p_d^{T_j}(\mathbf{x}_1) = 0, j \neq 1$). The generator can learn to generate these images e.g. $\mathbf{x}_1$ to achieve maximum of $\Phi(G,C^*)$, without actually learning the entire $P_d$.
  }
  \label{mixture_of_distribution}
   \vspace{-0.3cm}
\end{figure}

We analyze the SS tasks in \cite{chen-arxiv-2018} (Figure \ref{proposed_model}a).
We assume that all networks $D, G, C$ have enough capacity \cite{goodfellow-nisp-2014}. Refer to the Appendix \ref{appendix_a} for full derivation. Let $D^*$ and $C^*$ be the optimal discriminator and optimal classifier respectively at an equilibrium point. We assume that we have an optimal $D^*$ of the GAN task. We focus on $C^*$ of SS task. Let $p^{T_k}(\mathbf{x})$ be the probability of sample $\mathbf{x}$ 
under transformation by $T_k$
(Figure~\ref{mixture_of_distribution}). $p_d^{T_k}(\mathbf{x}), p_g^{T_k}(\mathbf{x})$ denotes the probability $p^{T_k}(\mathbf{x})$ of data sample ($\mathbf{x} \sim P_d^T$) or generated sample ($\mathbf{x} \sim P_g^T$) respectively.

\begin{proposition}
The optimal classifier $C^*$ of Eq. \ref{gan_dis_obj_ss} is:
\begin{equation}
C^*_k(\mathbf{x}) = \frac{p_d^{T_k}(\mathbf{x})}{\sum_{k=1}^{K}p_d^{T_k}(\mathbf{x})}
\label{optimal_c_d}
\end{equation}
\label{prop_1}
\end{proposition}
\noindent \textit{Proof}. Refer to our proof in Appendix \ref{appendix_a} for optimal $C^*$.

\begin{theorem}
Given optimal classifier $C^*$ for SS task $\Psi(C)$,
at the equilibrium point, maximizing SS task $\Phi(G,C^*)$ of Eq. \ref{gan_gen_obj_ss} is equal to maximizing:
\begin{equation}
\Phi(G,C^*) =  \frac{1}{K}\sum_{k=1}^{K} \bigg[\mathbb{E}_{\mathbf{x} \sim {P_g^{T_k}}}\log\Big(\frac{p_d^{T_k}(\mathbf{x})}{\sum_{k=1}^{K}p_d^{T_k}(\mathbf{x})}\Big)\bigg] 
= \frac{1}{K}\sum_{k=1}^{K}\mathcal{V}_{\Phi}^{T_k}(\mathbf{x})
\label{g_obj_jsd_1}
\end{equation}
\label{theorem_1}
\end{theorem}
\noindent \textit{Proof}. Refer to our proof in Appendix \ref{appendix_a}.\\

\begin{figure}
  \begin{minipage}[c]{0.26\textwidth}
          \begin{flushright}
          \includegraphics[width=3.51cm]{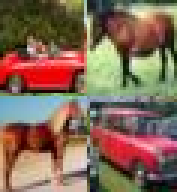}
          \end{flushright}
  \end{minipage}
  \begin{minipage}[c]{0.70\textwidth}
          \begin{center}
          \includegraphics[width=10.2cm]{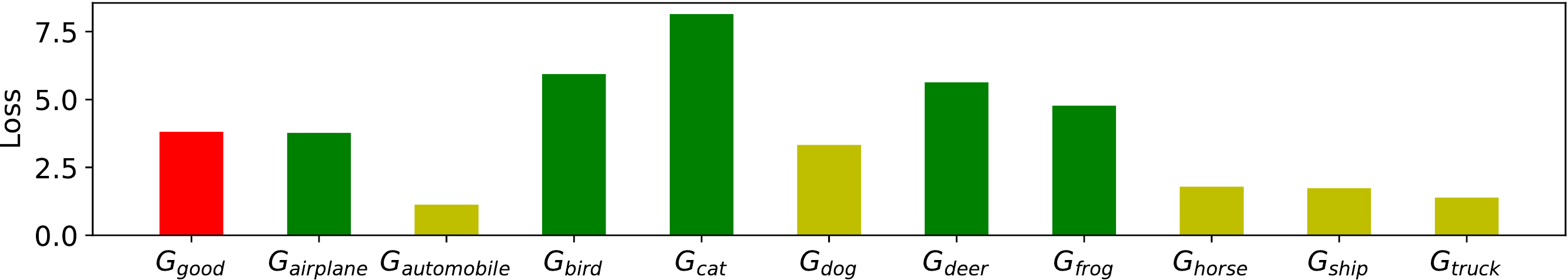}
          \vspace{-0.8cm}
          \end{center}
          \begin{center}
          \includegraphics[width=10.2cm]{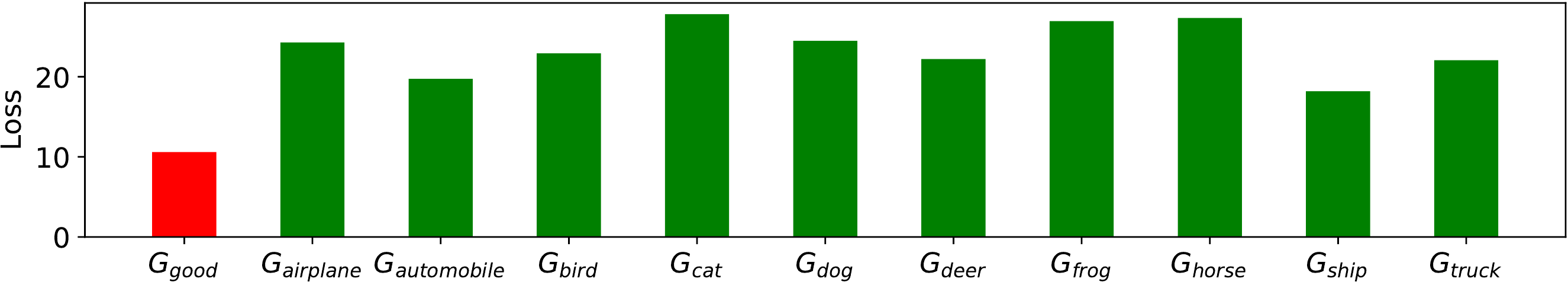}
          \end{center}
  \end{minipage}
  \caption{(a) Left: Example images that achieve minimal loss (or maximal $\Phi(G,C)$). For these images, rotation can be easily recognized: an image with a 90 degree rotated horse is likely due to applying $T_2$ rather than an original one. (b) Right (Top): the loss of original SS task, i.e. $-\Phi(G,C)$ computed over a good generator (red) and collapsed generators (green, yellow). 
  Some collapsed generators (e.g. one that generates only ``horse'') have smaller loss than the good generator under $-\Phi(G,C)$.
  (c) Right (Bottom): the loss of proposed MS task, $-\Phi^+(G, C)$, of a good generator (red) and collapsed generators (green). The good generator has the smallest loss under $-\Phi^+(G, C)$.
  }
  \label{classifiion_theorem_1}
\end{figure}

Theorem \ref{theorem_1} depicts learning of generator $G$
given the optimal $C^*$: selecting $G$ (hence $P_g$) to 
maximize $\Phi(G,C^*)$.
As $C^*$ is trained on real data,  $\Phi(G,C^*)$ encourages  $G$ to learn to generate realistic samples. However, we argue that $G$ can maximize $\Phi(G,C^*)$ without actually learning data distribution $P_d$. 
{\em In particular, it is sufficient for $G$ to  maximize $\Phi(G,C^*)$ by  simply learning to produce images which rotated version is rare (near zero probability).} Some example images are shown in Figure~\ref{classifiion_theorem_1}a. Intuitively, for these images, rotation can be easily recognized.

The argument can be developed from Theorem \ref{theorem_1}. From (\ref{g_obj_jsd_1}),
it can be shown that
$\mathcal{V}_{\Phi}^{T_k}(\mathbf{x}) \leq 0$ ($p_g^{T_k}(\mathbf{x}) >= 0$ and $\frac{p_d^{T_k}(\mathbf{x})}{\sum_{k=1}^{K}p_d^{T_k}(\mathbf{x})} \leq 1$).
One way for $G$ to achieve the maximum is to generate $\mathbf{x}$
such that 
$p_d^{T_1}(\mathbf{x}) \neq 0$ and 
$p_d^{T_j}(\mathbf{x}) = 0, j \neq 1$.
For these $\mathbf{x}$, the maximum 
$\mathcal{V}_{\Phi}^{T_k}(\mathbf{x}) = 0$ is attained. Note that $T_1$ corresponds to 0 degree rotation, i.e., no rotation.
Recall that 
$p_d^{T_k}(\mathbf{x})$ is the probability distribution of transformed data by $T_k$.
Therefore the condition 
$p_d^{T_1}(\mathbf{x}) \neq 0$ and $p_d^{T_j}(\mathbf{x}) = 0, j \neq 1$
means that there is no other rotated image resembling  $\mathbf{x}$, or equivalently, rotated $\mathbf{x}$ does not resemble any other images (Figure~\ref{mixture_of_distribution}). Therefore, the generator can exploit this ``loophole'' to maximize $\Phi(G,C^*)$ without actually learning the data distribution. 
In particular, even a mode-collapsed generator can achieve the maximum of $\Phi(G,C^*)$ by generating such images.

{\bf Empirical evidence.}
Empirically, our experiments (in Appendix \ref{ss_task_generator_learning}) show that  the FID of the models when using 
$\Phi(G,C)$
is poor except for very small
$\lambda_g$.
We further illustrate this issue by a toy empirical example using CIFAR-10.
We augment the  training images $\mathbf{x}$ with transformation data $T_k(\mathbf{x})$ to train the classifier $C$ to predict 
the rotation applied to $\mathbf{x}$.
This is the SS task of discriminator in 
Figure \ref{proposed_model}a.
Given this classifier $C$, we simulate the SS task of generator learning as follows.
To simulate the output of a good generator $G_{good}$ which generates diverse realistic samples, we choose the full test set of CIFAR-10 (10 classes) images and compute the cross-entropy loss, i.e. $-\Phi(G,C)$, when they are fed into $C$. 
To simulate the output of a mode-collapsed generator $G_{collapsed}$, we select samples from {\em one} class, e.g. ``horse'', and compute the cross-entropy loss when they are fed into $C$.
Fig. \ref{classifiion_theorem_1}b show that some $G_{collapsed}$ can outperform $G_{good}$ and achieve a smaller $-\Phi(G,C)$. E.g. a $G_{collapsed}$ that produces {\em only} ``horse'' samples outperform $G_{good}$ under $\Phi(G,C)$.  This example illustrates that, while 
$\Phi(G,C)$ may help the generator to create more realistic samples, it does not help the generator to prevent mode collapse. {\em In fact, as part of the multi-task loss
(see (\ref{gan_gen_obj_ss})), $\Phi(G,C)$ would undermine the learning of synthesizing diverse samples in the GAN task $\mathcal{V}(D,G)$.}

\section{Proposed method}
\label{proposed_method}

\subsection{Auxiliary Self-Supervised Tasks with Multi-class Minimax Game}

In this section, we propose improved SS tasks to address the issue (Fig. \ref{proposed_model}b).
Based on a multi-class minimax game, our classifier learns to distinguish the rotated samples from real data versus those from generated data. 
Our proposed SS tasks are  
$\Psi^+(G,C)$ and $\Phi^+(G, C)$ in (\ref{gan_dis_obj_ss_adv}) and (\ref{gan_gen_obj_ss_adv}) respectively.
Our discriminator objective is: 
\begin{equation}
\max_{D,C}\mathcal{V}(D,C,G) = \mathcal{V}(D,G) + \lambda_d \underbrace{\bigg(\mathbb{E}_{\mathbf{x} \sim {P_d^T}}\mathbb{E}_{T_k \sim \mathcal{T}}\log\Big(C_k(\mathbf{x})\Big) + \mathbb{E}_{\mathbf{x} \sim {P_g^T}}\mathbb{E}_{T_k \sim \mathcal{T}}\log\Big(C_{K+1}(\mathbf{x})\Big)\bigg)}_{\Psi^+(G,C)}
\label{gan_dis_obj_ss_adv}
\end{equation}
Eq. \ref{gan_dis_obj_ss_adv} means that we simultaneously distinguish generated samples, as the $(K + 1)$-th class, from the rotated real sample classes. Here, $C_{K+1}(\mathbf{x})$ is the $(K+1)$-th output for the fake class of classifier $C$. 

While rotated real samples are fixed samples that help prevent the classifier (discriminator) from forgetting, the class $K+1$ serves as the connecting point between generator and classifier, and the generator can directly challenge the classifier. Our technique resembles the original GAN by Goodfellow et al. \cite{goodfellow-nisp-2014}, but we generalize it for multi-class minimax game. Our generator objective is:
\begin{equation}
\min_G\mathcal{V}(D,C,G) = \mathcal{V}(D,G) - \lambda_g \underbrace{\bigg(\mathbb{E}_{\mathbf{x} \sim {P_g^T}}\mathbb{E}_{T_k \sim \mathcal{T}}\log\Big(C_k(\mathbf{x})\Big) - \mathbb{E}_{\mathbf{x} \sim P_g^T}\mathbb{E}_{T_k \sim \mathcal{T}}\log\Big(C_{K+1}(\mathbf{x})\Big)\bigg)}_{\Phi^+(G,C)}
\label{gan_gen_obj_ss_adv}
\end{equation}
$\Psi^+(G,C)$ and $\Phi^+(G,C)$ form a multi-class minimax game.
Note that, when we mention multi-class minimax game (or multi-class adversarial training), we refer to the SS tasks. The game for GAN task is the original by Goodfellow et al. \cite{goodfellow-nisp-2014}.  

\subsubsection{Theoretical Analysis}

\begin{proposition}
For fixed generator $G$, the optimal solution $C^*$ under Eq. \ref{gan_dis_obj_ss_adv} is:
\begin{equation}
C^*_k(\mathbf{x}) = \frac{p_d^T(\mathbf{x})}{p_g^T(\mathbf{x})}\frac{p_d^{T_k}(\mathbf{x})}{\sum_{k=1}^{K}p_d^{T_k}(\mathbf{x})} C^*_{K+1}(\mathbf{x})
\label{optimal_c_d_adv}
\end{equation} 
where $p_d^T(\mathbf{x})$ and $p_g^T(\mathbf{x})$ are probability of sample $\mathbf{x}$ in the mixture distributions $P_d^T$ and $P_g^T$ respectively.
\label{prop_2}
\end{proposition}

\noindent \textit{Proof}. Refer to our proof in Appendix \ref{appendix_a} for optimal $C^*$.

\begin{theorem}
Given optimal classifier $C^*$ obtained from multi-class minimax training $\Psi^+(G,C)$, at the equilibrium point, maximizing $\Phi^+(G,C^*)$ is equal to maximizing Eq. \ref{g_obj_jsd_2}:
\begin{equation}
\Phi^+(G,C^*) = - \frac{1}{K}\bigg[\sum_{k=1}^{K}\mathrm{KL}(P_g^{T_k}||P_d^{T_k})\bigg] + \frac{1}{K}\sum_{k=1}^{K}\bigg[\mathbb{E}_{\mathbf{x} \sim {P_g^{T_k}}}\log\Big(\frac{p_d^{T_k}(\mathbf{x})}{\sum_{k=1}^{K}p_d^{T_k}(\mathbf{x})}\Big)\bigg]
\label{g_obj_jsd_2}
\end{equation}
\label{theorem_2}
\end{theorem}

\noindent \textit{Proof}. Refer to our proof in Appendix \ref{appendix_a}.

Note that proposed SS task objective (\ref{g_obj_jsd_2}) is different from the original SS task objective (\ref{g_obj_jsd_1}) with the KL divergence term.
Furthermore, note that $\mathrm{KL}(P_g^{T_k}||P_d^{T_k}) = \mathrm{KL}(P_g||P_d)$, as rotation $T_k$ is an affine transform and KL divergence is invariant under affine transform (our proof in Appendix \ref{appendix_a}).
Therefore, the improvement is clear: {\em Proposed SS tasks $\Big(\Psi^+(.), \Phi^+(.)\Big)$ work together to improve the matching of $P_g$ and $P_d$ by leveraging the rotated samples.} For a given $P_g$, feedbacks are computed from not only  $\mathrm{KL}(P_g||P_d)$ but also $\mathrm{KL}(P_g^{T_k}||P_d^{T_k})$ via the rotated samples. Therefore, $G$ has more feedbacks to improve $P_g$. We investigate the improvement of our method on toy dataset as in Section \ref{analysis_on_auxiliary_ss}. The setup is the same, except that now we replace models/cost functions of $-\Phi(G,C)$ with our proposed ones $-\Phi^+(G,C)$ (the design of $G_{good}$ and $G_{collapsed}$ are the same). The loss now is shown in Fig. \ref{classifiion_theorem_1}c. Comparing Fig. \ref{classifiion_theorem_1}c and Fig. \ref{classifiion_theorem_1}b, the improvement using our proposed model can be observed: $G_{good}$ has the lowest loss under our proposed model. Note that, since optimizing KL divergence is not easy because it is asymmetric and could be biased to one direction \cite{tu-nips-2017}, in our implementation, we use a slightly modified version as described in the Appendix. 

\section{Experiments}

We measure the diversity and quality of generated samples via the Fr\'echet Inception Distance (FID) \cite{heusel-arxiv-2017}. FID is computed with 10K real samples and 5K generated samples exactly as in \cite{miyato-iclr-2018} if not precisely mentioned. 
We report the best FID attained in 300K iterations as in \cite{xiang-arxiv-2017, li-nips-2017, tran-eccv-2018, yazici-arxiv-2018}.
We integrate our proposed techniques into two baseline models (SSGAN \cite{chen-arxiv-2018} and Dist-GAN \cite{tran-eccv-2018}). We conduct experiments mainly on CIFAR-10 and STL-10 (resized into $48 \times 48$ as in \cite{miyato-iclr-2018}). We also provide additional experiments of CIFAR-100, Imagenet $32\times32$ and Stacked-MNIST.

For Dist-GAN \cite{tran-eccv-2018}, we evaluate three versions implemented with different network architectures: DCGAN architecture \cite{radford-arxiv-2015}, CNN architectures of SN-GAN \cite{miyato-iclr-2018} (referred as SN-GAN architecture) and ResNet architecture \cite{gulrajani-arxiv-2017}. We recall these network architectures in Appendix \ref{network-architecture}. We use ResNet architecture \cite{gulrajani-arxiv-2017} for experiments of CIFAR-100, Imagenet $32\times32$, and tiny K/4, K/2 architectures \cite{metz-arxiv-2016} for Stacked MNIST. We keep all parameters suggested in the original work and focus to understand the contribution of our proposed techniques. For SSGAN \cite{chen-arxiv-2018}, we use the ResNet architecture as implemented in the official code\footnote{https://github.com/google/compare\_gan}.

In our experiments, we use
{\bf SS}
to denote the original self-supervised tasks proposed in \cite{chen-arxiv-2018},  and we use {\bf MS} to denote our proposed self-supervised tasks ``Multi-class mini-max game based Self-supervised tasks". Details of the experimental setup and network parameters are discussed in Appendix \ref{appendix_b}.

We have conducted extensive experiments. Setup and results are discussed in Appendix \ref{appendix_b}.
In this section, we highlight the main results:
\begin{itemize}
  \item Comparison between {\bf SS} and our proposed {\bf MS} using the same baseline.
  \item Comparison between our proposed {\bf baseline + MS} and other state-of-the-art unconditional and conditional GAN. We emphasize that our proposed {\bf baseline + MS} is unconditional and does not use any label.
\end{itemize}

\subsection{Comparison between {\bf SS} and our proposed {\bf MS} using the same baseline}
\label{compare_ss_ms}
\begin{figure}
  \centering
  \includegraphics[width=3.3cm,keepaspectratio]{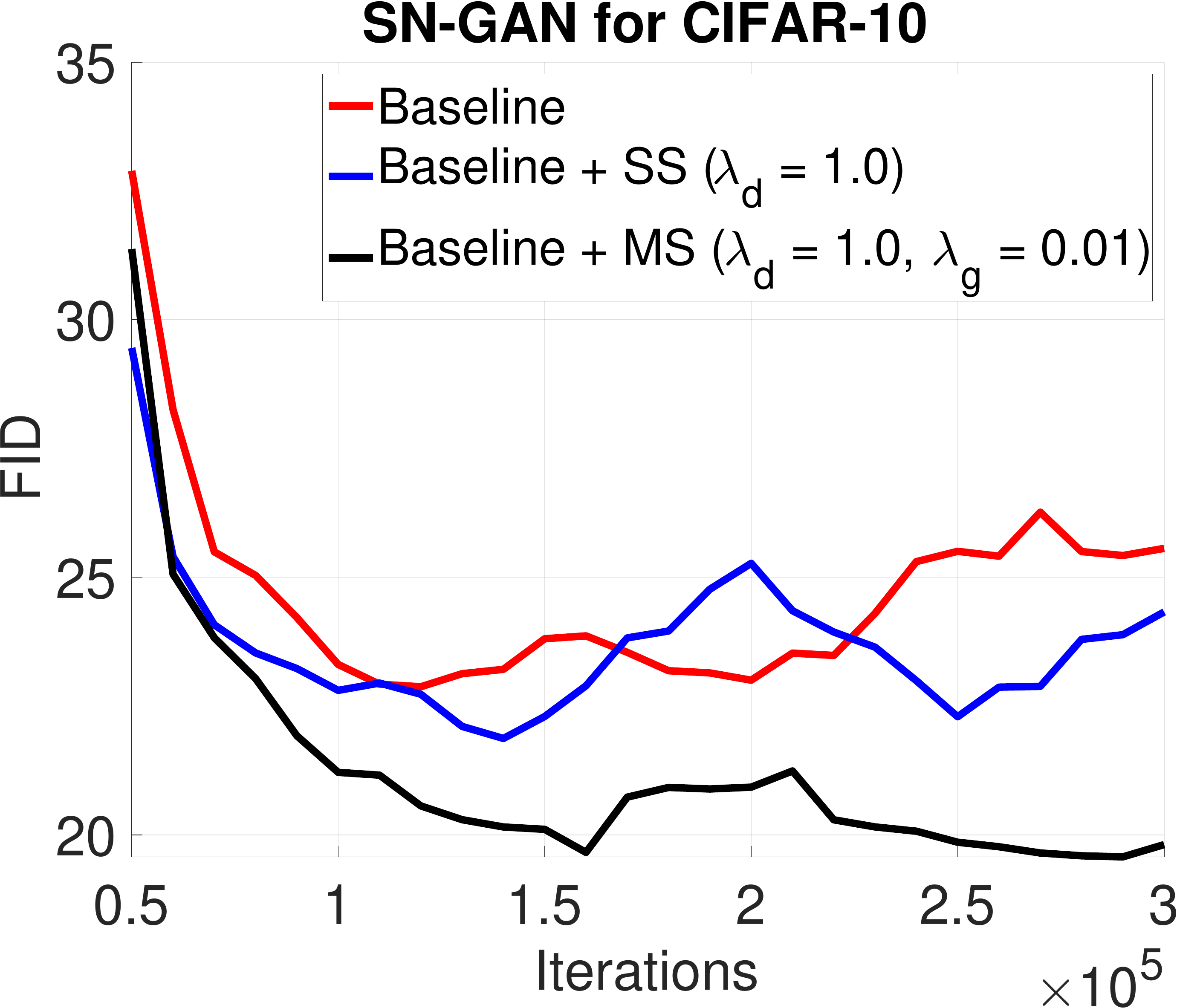}
  \includegraphics[width=3.3cm,keepaspectratio]{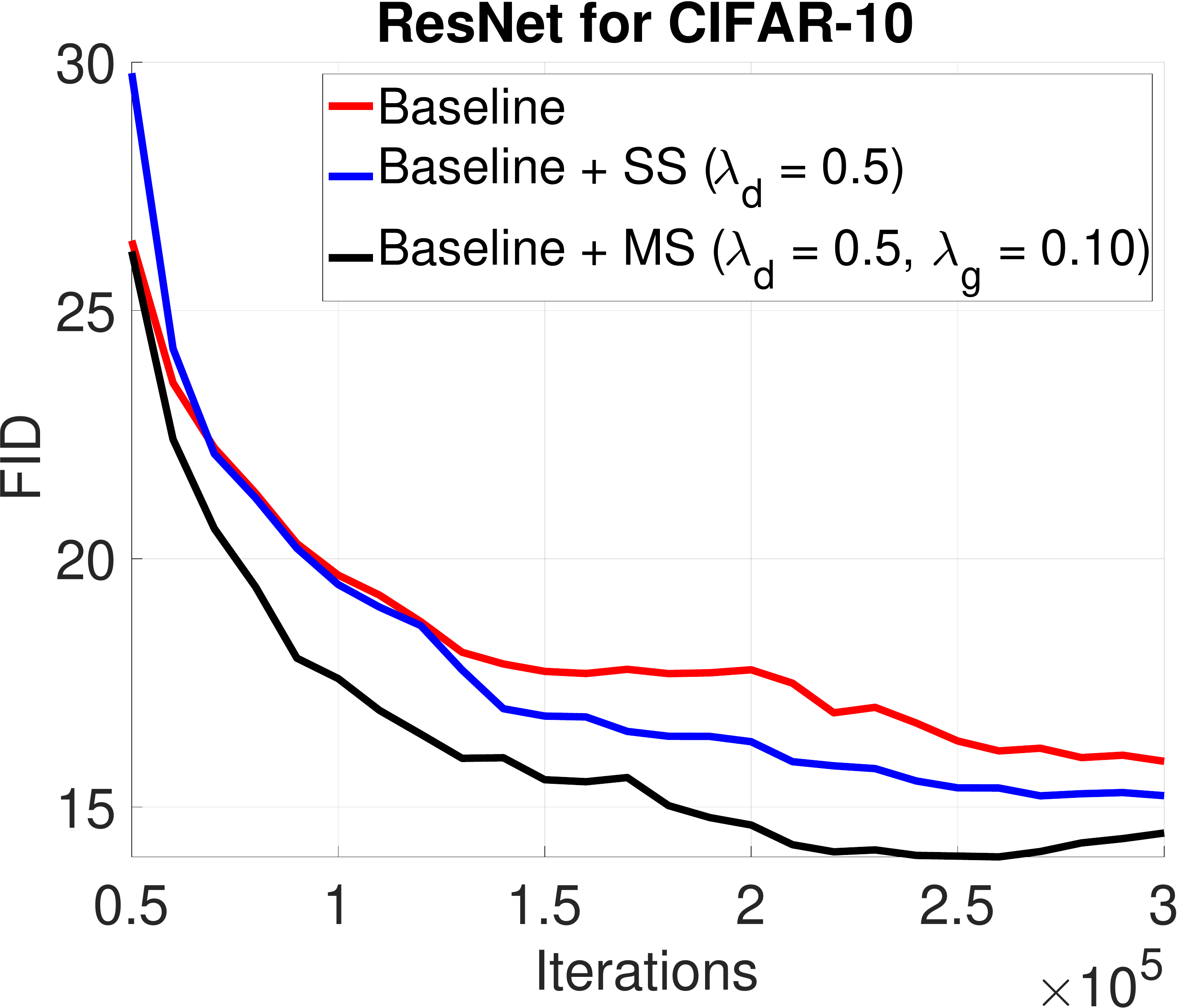}
  \includegraphics[width=3.3cm,keepaspectratio]{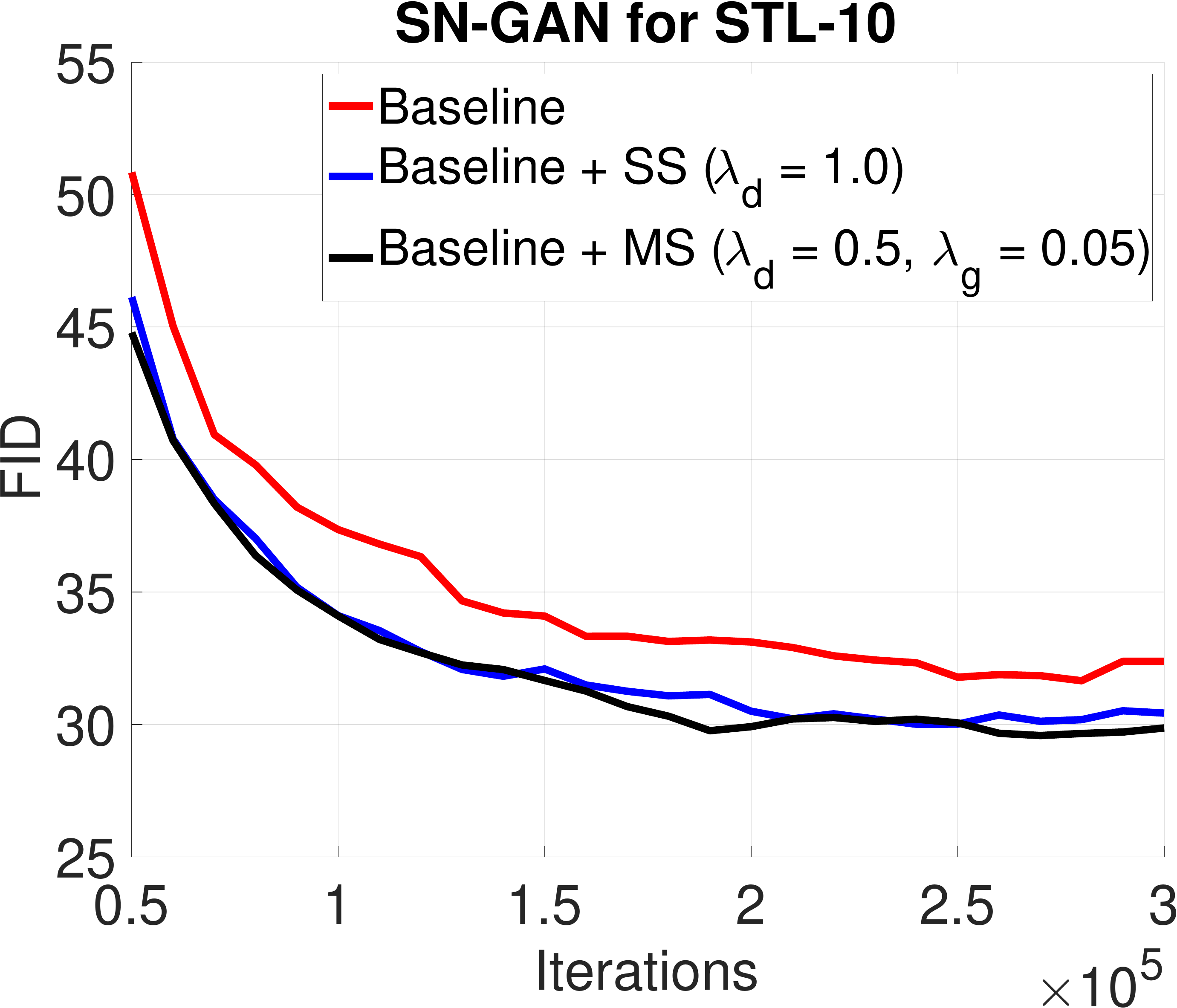}
  \includegraphics[width=3.3cm,keepaspectratio]{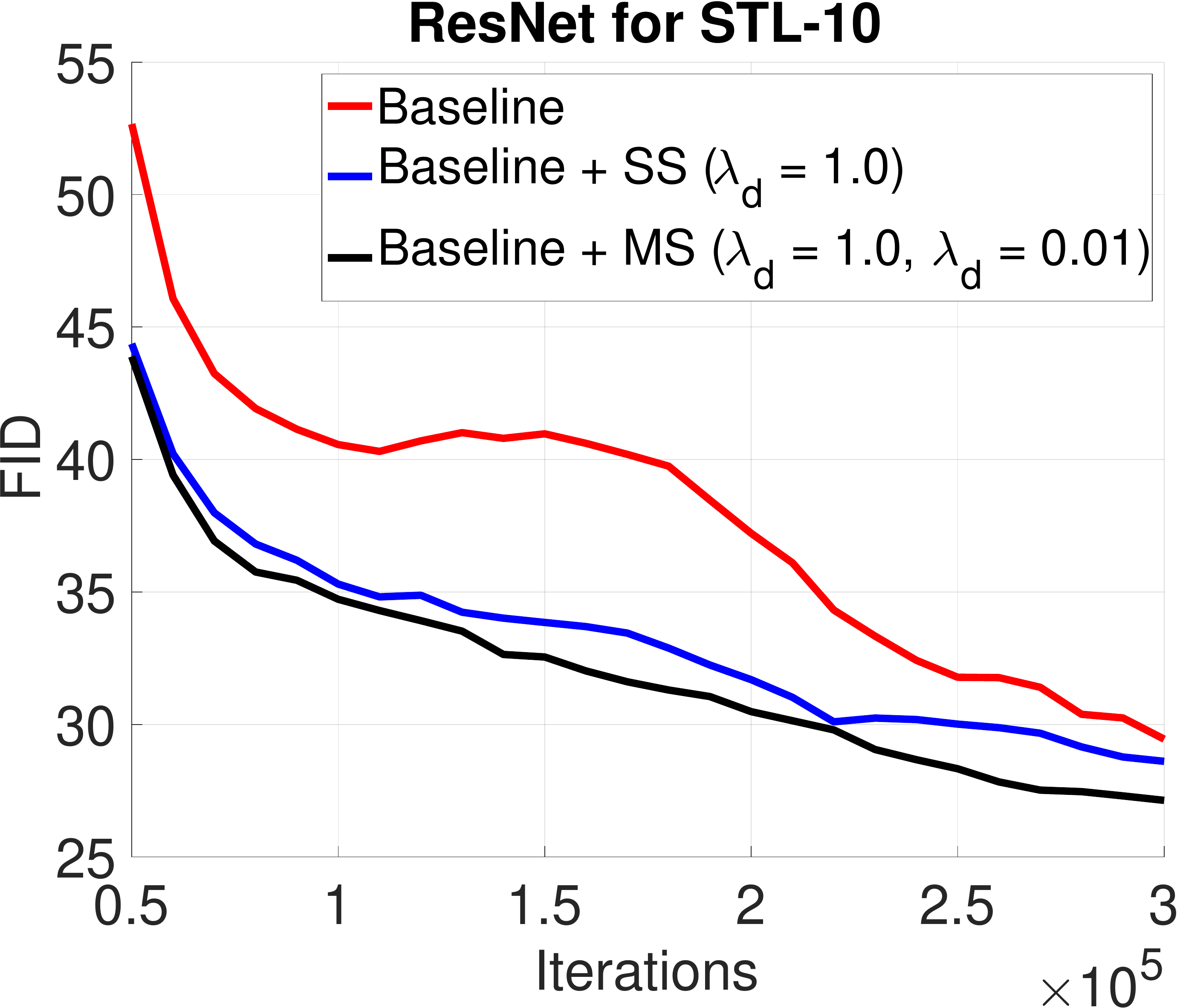}
  \caption{Compare {\bf SS} (original SS tasks proposed in \cite{chen-arxiv-2018}) and {\bf MS} (our proposed Multi-class mini-max game based Self-supervised tasks). The {\bf baseline} is Dist-GAN \cite{tran-eccv-2018}, implemented with {\bf SN-GAN} networks (CNN architectures in \cite{miyato-iclr-2018}) and {\bf ResNet}. Two datasets are used, CIFAR-10 and STL-10. For each experiment, we use the best $\lambda_d, \lambda_g$ for the models, obtained through extensive search (Appendix \ref{appendix_state_of_the_art}). Note that $\lambda_g = 0$ is the best for ``Baseline + SS'' in all experiments. The results suggest consistent improvement using our proposed self-supervised tasks.
  }
  \label{ss_d_g_finetuning_all}
\end{figure}

Results are shown in 
Fig. \ref{ss_d_g_finetuning_all} using Dist-GAN \cite{tran-eccv-2018} as the baseline.
For each experiment and for each approach ({\bf SS} or {\bf MS}), 
we obtain the best $\lambda_g$ and $\lambda_d$ using extensive search (see Appendix \ref{appendix_state_of_the_art} for details), and we use the best $\lambda_g$ and $\lambda_d$ in the comparison depicted in Fig. \ref{ss_d_g_finetuning_all}. 
In our experiments, we observe that Dist-GAN has stable convergence. Therefore, we use it in these experiments. As shown in Fig. \ref{ss_d_g_finetuning_all}, our proposed {\bf MS} outperforms the original {\bf SS} consistently.
More details can be found in Appendix \ref{appendix_state_of_the_art}.




\subsection{Comparison between our proposed method with other state-of-the-art GAN}

Main results are shown in Table \ref{state_of_the_art}. Details of this comparison can be found in Appendix \ref{appendix_state_of_the_art}. The best  $\lambda_g$ and $\lambda_d$ as in Figure \ref{ss_d_g_finetuning_all} are used in this comparison.
The best FID attained in 300K iterations are reported as in \cite{xiang-arxiv-2017, li-nips-2017, tran-eccv-2018, yazici-arxiv-2018}.
Note that SN-GAN method \cite{miyato-iclr-2018} attains the best FID at about 100K iterations with ResNet and it diverges afterward. Similar observation is also discussed in \cite{chen-arxiv-2018}. 

As shown in Table \ref{state_of_the_art}, our method (Dist-GAN + MS) consistently outperforms the baseline Dist-GAN and other state-of-the-art GAN. These results confirm the effectiveness of our proposed self-supervised tasks based on multi-class minimax game.

\begin{table}
  \small
  \caption{Comparison with other state-of-the-art GAN on CIFAR-10 and STL-10 datasets. 
  We report the best FID of the methods.
  Two network architectures are used: {\bf SN-GAN} networks (CNN architectures in \cite{miyato-iclr-2018}) and {\bf ResNet}.
  The FID scores are extracted from the respective papers when available.
  {\bf SS} denotes the original SS tasks proposed in \cite{chen-arxiv-2018}. {\bf MS} denotes our proposed self-supervised tasks. 
  `*': FID is computed with 10K-10K samples as in \cite{chen-arxiv-2018}. All compared GAN are unconditional, except SAGAN and BigGAN.
  SSGAN$^+$ is SS-GAN in \cite{chen-arxiv-2018} but using the best parameters we have obtained. 
  In SSGAN$^+$ + MS, we replace the original {\bf SS} in author's code with our proposed {\bf MS}.}

  \label{state_of_the_art}
  \centering
  \begin{tabular}{llllll}
    \toprule
    & \multicolumn{2}{c}{\textbf{SN-GAN}} & \multicolumn{2}{c}{\textbf{ResNet}} \\
    \cmidrule(r){2-3}  \cmidrule(r){4-6}
    \textbf{Methods}  & \textbf{CIFAR-10}     & \textbf{STL-10}     & \textbf{CIFAR-10}   & \textbf{STL-10}  & \textbf{CIFAR-10$ ^*$}\\
    \midrule
    GAN-GP \cite{miyato-iclr-2018}    			      & 37.7   & -      & - & - & - \\
    WGAN-GP \cite{miyato-iclr-2018}    			      & 40.2   & 55.1   & - & - & - \\
    SN-GAN \cite{miyato-iclr-2018}    			      & 25.5   & 43.2   & 21.70 $\pm$ .21 & 40.10 $\pm$ .50  & 19.73 \\
    SS-GAN \cite{chen-arxiv-2018}                     & -      & -      & -               & -               & 15.65 \\
    Dist-GAN \cite{tran-eccv-2018}  			      & 22.95  & 36.19  & 17.61 $\pm$ .30 & 28.50 $\pm$ .49  & 13.01 \\
    GN-GAN \cite{tran-aaai-2018}                      & 21.70  & 30.80  & 16.47 $\pm$ .28 & - & - \\
    \hline
    SAGAN \cite{zhang-arxiv-2018} (cond.)             & -  & -  & 13.4 (best) & - & - \\
    BigGAN \cite{brock-iclr-2018} (cond.)             & -  & -  & 14.73       & - & - \\
    \hline
    SSGAN$^+$                                         & -  & -  & -  & -  & 20.47 \\
    \textbf{Ours(SSGAN$^+$ + MS)}                     & -  & -  & -  & -  & 19.89 \\
    \hline

    Dist-GAN + SS                                     & 21.40  & 29.79 & 14.97 $\pm$ .29 & 27.98 $\pm$ .38 & 12.37 \\ 

    \textbf{Ours(Dist-GAN + MS)}                      & \textbf{18.88}  & \textbf{27.95} & \textbf{13.90 $\pm$ .22} & \textbf{27.10 $\pm$ .34} & \textbf{11.40} \\
    \bottomrule
  \end{tabular}
  \vspace{-0.4cm}
\end{table}

We have also extracted the FID reported in \cite{chen-arxiv-2018}, i.e. SSGAN with the original SS tasks proposed there. In  this case, we follow exactly their settings and compute FID using 10K real samples and 10K fake samples. Our model achieves better FID score than SSGAN with exactly the same ResNet architecture on CIFAR-10 dataset.  See results under the column CIFAR-10$^*$ in Table \ref{state_of_the_art}.

Note that we have tried to reproduce the results of SSGAN 
using its published code, but we were unable to achieve similar results as reported in the original paper \cite{chen-arxiv-2018}. 
We have performed extensive search and we use the obtained best parameter to report the results as 
SSGAN$^+$ in Table \ref{state_of_the_art} (i.e., SSGAN$^+$ uses the published code and the best parameters we obtained). 
We use this code and setup to 
compare 
{\bf SS}  and {\bf MS}, i.e. we replace the {\bf SS} code in the system with {\bf MS} code, and obtain 
``SSGAN$^+$ + MS''. As shown in Table \ref{state_of_the_art}, our ``SSGAN$^+$ + MS'' achieves better FID than  SSGAN$^+$. 
The improvement is consistent with Figure \ref{ss_d_g_finetuning_all} when Dist-GAN is used as the baseline.
More detailed experiments can be found in the Appendix. 
We have also compared SSGAN$^+$ and our system (SSGAN$^+$ + MS) on CelebA ($64 \times 64$). In this experiment, we use a small DCGAN architecture provided in the authors' code. Our proposed MS outperforms the original SS, with FID improved  
from $35.03$ to $33.47$. This experiment  again confirms the effectiveness of our proposed MS.

We conduct additional experiments on CIFAR-100 and ImageNet 32$\times$32 to compare \textbf{SS} and \textbf{MS} with Dist-GAN baseline. We use the same ResNet architecture as Section \ref{appendix_state_of_the_art} on CIFAR-10 for this study, and we use the best parameters $\lambda_d$ and $\lambda_g$ selected in Section \ref{appendix_state_of_the_art} for ResNet architecture. Experimental results in Table  \ref{new_experiments_cifar100_imagenet} show that our \textbf{MS} consistently outperform \textbf{SS} for  all benchmark datasets. For ImageNet 32$\times$32 we report the \textit{best} FID for \textbf{SS} because the model suffers serious mode collapse at the end of training. Our \textbf{MS} achieves the best performance at the end of training.

\begin{table}
    \small
    \centering
    \caption{Results on CIFAR-100 and ImageNet 32$\times$32. We use baseline model Dist-GAN with ResNet architecture. We  follow the same experiment setup as above. {\bf SS}: proposed in [4]; {\bf MS}: this work.}
    \begin{tabular}{ c  c  c }
    \toprule
    \textbf{Datasets} & \textbf{SS} & \textbf{MS} \\
    \hline
    CIFAR-100 (10K-5K FID)                    & 21.02 & 19.74  \\
    ImageNet 32$\times$32 (10K-10K FID)       & 17.1 & 12.3 \\
    \bottomrule
    \end{tabular}
    \label{new_experiments_cifar100_imagenet}
    \vspace{-0.4cm}
\end{table}

We also evaluate the diversity of our generator on Stacked MNIST \cite{metz-arxiv-2016}. Each image of this dataset is synthesized by stacking any three random MNIST digits. We follow exactly the same experiment setup with tiny architectures $K/4$, $K/2$ and evaluation protocol of \cite{metz-arxiv-2016}. We measure the quality of methods by the number of covered modes (higher is better) and KL divergence (lower is better). Refer to \cite{metz-arxiv-2016} for more details. Table. \ref{new_experiments_stacked_mnist} shows that our proposed \textbf{MS} outperforms \textbf{SS} for both mode number and KL divergence. Our approach  significantly outperforms state-of-the-art \cite{tran-eccv-2018,karras-iclr-2018}. The means and standard deviations of \textbf{MS} and \textbf{SS} are computed from eight runs (we re-train our GAN model from the scratch for each run). The results are reported with best $(\lambda_d,\lambda_g)$ of $\mathbf{MS}$: $(0.5, 0.2)$ for $K/4$ architecture and $(1.0, 1.0)$ for $K/2$ architecture. Similarly, best $(\lambda_d,\lambda_g)$ of $\mathbf{SS}$: $(0.5, 0.0)$ for $K/4$ architecture and $(1.0, 0.0)$ for $K/2$ architecture.

\begin{table}
    \centering
    \scriptsize
    \caption{Comparing to state-of-the-art methods on Stacked MNIST with tiny $K/4$ and $K/2$ architectures \cite{metz-arxiv-2016}. We also follow the same experiment setup of \cite{metz-arxiv-2016}. Baseline model: Dist-GAN. {\bf SS}: proposed in \cite{chen-arxiv-2018}; {\bf MS}: this work. Our method {\bf MS} achieves the best results for this dataset with both architectures, outperforming state-of-the-art \cite{tran-eccv-2018,karras-iclr-2018} by a significant margin.}
    \begin{tabular}{ c  c  c  c  c  c  c }
    \toprule
    \textbf{Arch} & \textbf{Unrolled GAN \cite{metz-arxiv-2016}} & \textbf{WGAN-GP \cite{gulrajani-arxiv-2017}} & \textbf{Dist-GAN \cite{tran-eccv-2018}} & \textbf{Pro-GAN \cite{karras-iclr-2018}} & \textbf{\cite{tran-eccv-2018}+SS} & \textbf{Ours(\cite{tran-eccv-2018}+MS)}\\ 
    \hline
    \textbf{K/4, \#} & 372.2 $\pm$ 20.7 & 640.1 $\pm$ 136.3 & 859.5 $\pm$ 68.7 & 859.5 $\pm$ 36.2 & 906.75 $\pm$ 26.15 & 926.75 $\pm$ 32.65 \\ 
    \textbf{K/4, KL} & 4.66 $\pm$ 0.46 & 1.97 $\pm$ 0.70 & 1.04 $\pm$ 0.29 & 1.05 $\pm$ 0.09 & 0.90 $\pm$ 0.13 & 0.78 $\pm$ 0.13\\
    \hline
    \textbf{K/2, \#} & 817.4 $\pm$ 39.9 & 772.4 $\pm$ 146.5 & 917.9 $\pm$ 69.6 & 919.8 $\pm$ 35.1 & 957.50 $\pm$ 31.23 & 976.00 $\pm$ 10.04\\ 
    \textbf{K/2, KL} & 1.43 $\pm$ 0.12 & 1.35 $\pm$ 0.55  & 1.06 $\pm$ 0.23 & 0.82 $\pm$ 0.13 & 0.61 $\pm$ 0.15 & 0.52 $\pm$ 0.07\\
    \bottomrule
    \end{tabular}
    \label{new_experiments_stacked_mnist}
    \vspace{-0.4cm}
\end{table}

Finally, in Table \ref{state_of_the_art}, we compare our FID to SAGAN \cite{zhang-arxiv-2018} (a state-of-the-art conditional GAN) and BigGAN \cite{brock-iclr-2018}. We perform the experiments under the same conditions using ResNet architecture on the  CIFAR-10 dataset. We report the best FID that SAGAN can achieve. As SAGAN paper does not have CIFAR-10 results \cite{zhang-arxiv-2018}, we run the published SAGAN code and select the best parameters to obtain the results for CIFAR-10. For BigGAN, we extract best FID from original paper. Although our method is unconditional, our best FID is very close  to that of these state-of-the-art conditional GAN. This validates the effectiveness of our design. 
Generated images using our system can be found in 
Figures \ref{cifar_resnet_samples} and \ref{stl_resnet_samples} of Appendix \ref{appendix_b}.

\section{Conclusion}

We provide theoretical and empirical analysis on auxiliary self-supervised task for GAN.
Our analysis reveals
the limitation of the existing work. To address the limitation, we propose multi-class minimax game based self-supervised tasks.
Our 
proposed self-supervised tasks leverage the rotated samples to provide better feedback in matching the data and generator distributions.
Our theoretical and empirical analysis support improved convergence of our design.
Our proposed SS tasks can be easily incorporated into existing GAN models. Experiment results suggest that they help boost the performance of baseline implemented with various network architectures on the CIFAR-10, CIFAR-100, STL-10, CelebA, Imagenet $32\times32$, and Stacked-MNIST datasets. The best version of our proposed method establishes state-of-the-art FID scores on all these benchmark datasets.

%

\pagebreak

\subsubsection*{Acknowledgements}

This work was  supported by ST Electronics and the National Research Foundation(NRF), Prime Minister's Office, Singapore under Corporate Laboratory @ University Scheme (Programme Title: STEE Infosec - SUTD Corporate Laboratory). This research was also supported by the National Research Foundation Singapore under its AI Singapore Programme [Award Number: AISG-100E-2018-005].
This research was also supported in part by the Energy Market Authority (EP award no. NRF2017EWT-EP003-061).
This project was  also supported by SUTD project PIE-SGP-AI-2018-01.

{\small
\bibliographystyle{plain} 
\bibliography{biblio}

\begin{thebibliography}{10}

\bibitem{arjovsky-arxiv-2017a}
Martin Arjovsky and L{\'e}on Bottou.
\newblock Towards principled methods for training generative adversarial
  networks.
\newblock {\em arXiv preprint arXiv:1701.04862}, 2017.

\bibitem{brock-iclr-2018}
Andrew Brock, Jeff Donahue, and Karen Simonyan.
\newblock Large scale gan training for high fidelity natural image synthesis.
\newblock {\em arXiv preprint arXiv:1809.11096}, 2018.

\bibitem{che-arxiv-2016}
Tong Che, Yanran Li, Athul~Paul Jacob, Yoshua Bengio, and Wenjie Li.
\newblock Mode regularized generative adversarial networks.
\newblock {\em CoRR}, 2016.

\bibitem{chen-arxiv-2018}
Ting Chen, Xiaohua Zhai, Marvin Ritter, Mario Lucic, and Neil Houlsby.
\newblock Self-supervised gans via auxiliary rotation loss.
\newblock In {\em CVPR}, 2019.

\bibitem{chen-arxiv-2016}
Xi~Chen, Yan Duan, Rein Houthooft, John Schulman, Ilya Sutskever, and Pieter
  Abbeel.
\newblock Infogan: Interpretable representation learning by information
  maximizing generative adversarial nets.
\newblock In {\em Advances in Neural Information Processing Systems}, pages
  2172--2180, 2016.

\bibitem{doersch-cvpr-2015}
Carl Doersch, Abhinav Gupta, and Alexei~A Efros.
\newblock Unsupervised visual representation learning by context prediction.
\newblock In {\em CVPR}, 2015.

\bibitem{donahue-arxiv-2016}
Jeff Donahue, Philipp Kr{\"a}henb{\"u}hl, and Trevor Darrell.
\newblock Adversarial feature learning.
\newblock {\em arXiv preprint arXiv:1605.09782}, 2016.

\bibitem{dumoulin-arxiv-2016}
Vincent Dumoulin, Ishmael Belghazi, Ben Poole, Alex Lamb, Martin Arjovsky,
  Olivier Mastropietro, and Aaron Courville.
\newblock Adversarially learned inference.
\newblock {\em arXiv preprint arXiv:1606.00704}, 2016.

\bibitem{ghosh-cvpr-2018}
Arnab Ghosh, Viveka Kulharia, Vinay~P Namboodiri, Philip~HS Torr, and Puneet~K
  Dokania.
\newblock Multi-agent diverse generative adversarial networks.
\newblock In {\em CVPR}, 2018.

\bibitem{gidaris-iclr-2018}
Spyros Gidaris, Praveer Singh, and Nikos Komodakis.
\newblock Unsupervised representation learning by predicting image rotations.
\newblock {\em ICLR}, 2018.

\bibitem{goodfellow-nips-2016}
Ian Goodfellow.
\newblock Nips 2016 tutorial: Generative adversarial networks.
\newblock {\em arXiv preprint arXiv:1701.00160}, 2016.

\bibitem{goodfellow-nisp-2014}
Ian Goodfellow, Jean Pouget-Abadie, Mehdi Mirza, Bing Xu, David Warde-Farley,
  Sherjil Ozair, Aaron Courville, and Yoshua Bengio.
\newblock Generative adversarial nets.
\newblock In {\em NIPS}, pages 2672--2680, 2014.

\bibitem{gulrajani-arxiv-2017}
Ishaan Gulrajani, Faruk Ahmed, Martin Arjovsky, Vincent Dumoulin, and Aaron~C
  Courville.
\newblock Improved training of wasserstein gans.
\newblock In {\em Advances in Neural Information Processing Systems}, pages
  5767--5777, 2017.

\bibitem{heusel-arxiv-2017}
Martin Heusel, Hubert Ramsauer, Thomas Unterthiner, Bernhard Nessler, and Sepp
  Hochreiter.
\newblock Gans trained by a two time-scale update rule converge to a local nash
  equilibrium.
\newblock In {\em Advances in Neural Information Processing Systems}, pages
  6626--6637, 2017.

\bibitem{hoang-arxiv-2018}
Quan Hoang, Tu~Dinh Nguyen, Trung Le, and Dinh Phung.
\newblock Mgan: Training generative adversarial nets with multiple generators.
\newblock 2018.

\bibitem{isola-cvpr-2017}
Phillip Isola, Jun-Yan Zhu, Tinghui Zhou, and Alexei~A Efros.
\newblock Image-to-image translation with conditional adversarial networks.
\newblock {\em CVPR}, 2017.

\bibitem{karras-iclr-2018}
Tero Karras, Timo Aila, Samuli Laine, and Jaakko Lehtinen.
\newblock Progressive growing of gans for improved quality, stability, and
  variation.
\newblock {\em arXiv preprint arXiv:1710.10196}, 2017.

\bibitem{karras-cvpr-2019}
Tero Karras, Samuli Laine, and Timo Aila.
\newblock A style-based generator architecture for generative adversarial
  networks.
\newblock In {\em CVPR}, 2019.

\bibitem{kingma-arxiv-2013}
Diederik~P Kingma and Max Welling.
\newblock Auto-encoding variational bayes.
\newblock {\em arXiv preprint arXiv:1312.6114}, 2013.

\bibitem{kirkpatrick-2017-nas}
James Kirkpatrick, Razvan Pascanu, Neil Rabinowitz, Joel Veness, Guillaume
  Desjardins, Andrei~A Rusu, Kieran Milan, John Quan, Tiago Ramalho, Agnieszka
  Grabska-Barwinska, et~al.
\newblock Overcoming catastrophic forgetting in neural networks.
\newblock {\em Proceedings of the national academy of sciences}, 2017.

\bibitem{kodali-arxiv-2017}
Naveen Kodali, Jacob Abernethy, James Hays, and Zsolt Kira.
\newblock On convergence and stability of gans.
\newblock {\em arXiv preprint arXiv:1705.07215}, 2017.

\bibitem{larsen-arxiv-2015}
Anders Boesen~Lindbo Larsen, S{\o}ren~Kaae S{\o}nderby, Hugo Larochelle, and
  Ole Winther.
\newblock Autoencoding beyond pixels using a learned similarity metric.
\newblock {\em arXiv preprint arXiv:1512.09300}, 2015.

\bibitem{ledig-cvpr-2017}
Christian Ledig, Lucas Theis, Ferenc Husz{\'a}r, Jose Caballero, Andrew
  Cunningham, Alejandro Acosta, Andrew Aitken, Alykhan Tejani, Johannes Totz,
  Zehan Wang, et~al.
\newblock Photo-realistic single image super-resolution using a generative
  adversarial network.
\newblock In {\em CVPR}, 2017.

\bibitem{kwotsin:2019}
Kwot~Sin Lee, Ngoc-Trung Tran, and Ngai-Man Cheung.
\newblock Infomax-gan: Mutual information maximization for improved adversarial
  image generation.
\newblock In {\em NeurIPS 2019 Workshop on Information Theory and Machine
  Learning}, 2019.

\bibitem{li-nips-2017}
Chun-Liang Li, Wei-Cheng Chang, Yu~Cheng, Yiming Yang, and Barnab{\'a}s
  P{\'o}czos.
\newblock Mmd gan: Towards deeper understanding of moment matching network.
\newblock In {\em NIPS}, 2017.

\bibitem{lim-icdm-2018}
Swee~Kiat Lim, Yi~Loo, Ngoc-Trung Tran, Ngai-Man Cheung, Gemma Roig, and Yuval
  Elovici.
\newblock Doping: Generative data augmentation for unsupervised anomaly
  detection.
\newblock In {\em Proceeding of IEEE International Conference on Data Mining
  (ICDM)}, 2018.

\bibitem{liu-arxiv-2018}
Kanglin Liu.
\newblock Varying k-lipschitz constraint for generative adversarial networks.
\newblock {\em arXiv preprint arXiv:1803.06107}, 2018.

\bibitem{liu-cvpr-2019}
Xuanqing Liu and Cho-Jui Hsieh.
\newblock Rob-gan: Generator, discriminator and adversarial attacker.
\newblock In {\em CVPR}, 2019.

\bibitem{makhzani-arxiv-2015}
Alireza Makhzani, Jonathon Shlens, Navdeep Jaitly, and Ian Goodfellow.
\newblock Adversarial autoencoders.
\newblock In {\em International Conference on Learning Representations}, 2016.

\bibitem{metz-arxiv-2016}
Luke Metz, Ben Poole, David Pfau, and Jascha Sohl-Dickstein.
\newblock Unrolled generative adversarial networks.
\newblock {\em ICLR}, 2017.

\bibitem{miyato-iclr-2018}
Takeru Miyato, Toshiki Kataoka, Masanori Koyama, and Yuichi Yoshida.
\newblock Spectral normalization for generative adversarial networks.
\newblock {\em ICLR}, 2018.

\bibitem{tu-nips-2017}
Tu~Nguyen, Trung Le, Hung Vu, and Dinh Phung.
\newblock Dual discriminator generative adversarial nets.
\newblock In {\em NIPS}, 2017.

\bibitem{noroozi-iccv-2017}
Mehdi Noroozi, Hamed Pirsiavash, and Paolo Favaro.
\newblock Representation learning by learning to count.
\newblock In {\em ICCV}, 2017.

\bibitem{odena-icml-2017}
Augustus Odena, Christopher Olah, and Jonathon Shlens.
\newblock Conditional image synthesis with auxiliary classifier {GAN}s.
\newblock In {\em ICML}, 2017.

\bibitem{pathak-cvpr-2016}
Deepak Pathak, Philipp Krahenbuhl, Jeff Donahue, Trevor Darrell, and Alexei~A
  Efros.
\newblock Context encoders: Feature learning by inpainting.
\newblock In {\em CVPR}, 2016.

\bibitem{petzka-arxiv-2017}
Henning Petzka, Asja Fischer, and Denis Lukovnicov.
\newblock On the regularization of wasserstein gans.
\newblock {\em arXiv preprint arXiv:1709.08894}, 2017.

\bibitem{radford-arxiv-2015}
Alec Radford, Luke Metz, and Soumith Chintala.
\newblock Unsupervised representation learning with deep convolutional
  generative adversarial networks.
\newblock {\em arXiv preprint arXiv:1511.06434}, 2015.

\bibitem{reed-arxiv-2016}
Scott Reed, Zeynep Akata, Xinchen Yan, Lajanugen Logeswaran, Bernt Schiele, and
  Honglak Lee.
\newblock Generative adversarial text to image synthesis.
\newblock {\em arXiv preprint arXiv:1605.05396}, 2016.

\bibitem{roth-nips-2017}
Kevin Roth, Aurelien Lucchi, Sebastian Nowozin, and Thomas Hofmann.
\newblock Stabilizing training of generative adversarial networks through
  regularization.
\newblock In {\em Advances in Neural Information Processing Systems}, pages
  2018--2028, 2017.

\bibitem{salimans-nisp-2016}
Tim Salimans, Ian Goodfellow, Wojciech Zaremba, Vicki Cheung, Alec Radford, and
  Xi~Chen.
\newblock Improved techniques for training gans.
\newblock In {\em NIPS}, pages 2234--2242, 2016.

\bibitem{schlegl-ipmi-2017}
Thomas Schlegl, Philipp Seeb{\"{o}}ck, Sebastian~M. Waldstein, Ursula
  Schmidt{-}Erfurth, and Georg Langs.
\newblock Unsupervised anomaly detection with generative adversarial networks
  to guide marker discovery.
\newblock {\em CoRR}, abs/1703.05921, 2017.

\bibitem{tran-eccv-2018}
Ngoc-Trung Tran, Tuan-Anh Bui, and Ngai-Man Cheung.
\newblock Dist-gan: An improved gan using distance constraints.
\newblock In {\em ECCV}, 2018.

\bibitem{tran-aaai-2018}
Ngoc{-}Trung Tran, Tuan{-}Anh Bui, and Ngai{-}Man Cheung.
\newblock Improving gan with neighbors embedding and gradient matching.
\newblock In {\em AAAI}, 2019.

\bibitem{tran-arxiv-2019}
Ngoc-Trung Tran, Viet-Hung Tran, Ngoc-Bao Nguyen, and Ngai-Man Cheung.
\newblock An improved self-supervised gan via adversarial training.
\newblock {\em arXiv preprint arXiv:1905.05469}, 2019.

\bibitem{xiang-arxiv-2017}
Sitao Xiang and Hao Li.
\newblock On the effects of batch and weight normalization in generative
  adversarial networks.
\newblock {\em arXiv preprint arXiv:1704.03971}, 2017.

\bibitem{Yang_2019_ICCV}
Linxiao Yang, Ngai-Man Cheung, Jiaying Li, and Jun Fang.
\newblock Deep clustering by gaussian mixture variational autoencoders with
  graph embedding.
\newblock In {\em The IEEE International Conference on Computer Vision (ICCV)},
  October 2019.

\bibitem{yazici-arxiv-2018}
Yasin Yaz{\i}c{\i}, Chuan-Sheng Foo, Stefan Winkler, Kim-Hui Yap, Georgios
  Piliouras, and Vijay Chandrasekhar.
\newblock The unusual effectiveness of averaging in gan training.
\newblock {\em arXiv preprint arXiv:1806.04498}, 2018.

\bibitem{zenke-arxiv-2017}
Friedemann Zenke, Ben Poole, and Surya Ganguli.
\newblock Continual learning through synaptic intelligence.
\newblock {\em arXiv preprint arXiv:1703.04200}, 2017.

\bibitem{zhang-arxiv-2018}
Han Zhang, Ian Goodfellow, Dimitris Metaxas, and Augustus Odena.
\newblock Self-attention generative adversarial networks.
\newblock {\em arXiv preprint arXiv:1805.08318}, 2018.

\bibitem{zhang2-cvpr-2017}
Han Zhang, Tao Xu, Hongsheng Li, Shaoting Zhang, Xiaogang Wang, Xiaolei Huang,
  and Dimitris~N Metaxas.
\newblock Stackgan: Text to photo-realistic image synthesis with stacked
  generative adversarial networks.
\newblock In {\em CVPR}, 2017.

\bibitem{zhang-eccv-2016}
Richard Zhang, Phillip Isola, and Alexei~A Efros.
\newblock Colorful image colorization.
\newblock In {\em ECCV}, 2016.

\bibitem{zhang1-cvpr-2017}
Richard Zhang, Phillip Isola, and Alexei~A Efros.
\newblock Split-brain autoencoders: Unsupervised learning by cross-channel
  prediction.
\newblock In {\em CVPR}, 2017.

\bibitem{zhu-cvpr-2017}
Jun-Yan Zhu, Taesung Park, Phillip Isola, and Alexei~A Efros.
\newblock Unpaired image-to-image translation using cycle-consistent
  adversarial networkss.
\newblock In {\em ICCV}, 2017.

\end{thebibliography}
}


\clearpage
\appendix

In our paper, we perform an in-depth analysis to understand how SS tasks interact with the learning of the generator. We analyze the issues of SS tasks and propose to improve it with a multi-class minimax game. 
In this Appendix section, we provide detail information about our proofs, discussion, ablation study, network parameters and network architectures of models.

\section{Appendix: Proofs for Sections \ref{analysis_on_auxiliary_ss} and \ref{proposed_method}}
\label{appendix_a}

\noindent \textbf{Proposition \ref{prop_1}} (Proof.)

Let $T_k$ be the $k$-th type of transformation, and let $P_d^T$ be the distribution of the transformed real sample.
This section shows the proof for optimal $C^*$. $C_k(.)$ is the $k$-th soft-max output of $C$, hence $\sum_{k=1}^{K}C_k(\mathbf{x}) = 1, \forall \mathbf{x}$. $\Psi(C)$ can be re-written as: 

\begin{equation}
\Psi(C) = \int {p_d^{T}}(\mathbf{x}) \bigg( \sum_{k=1}^{K}p({T_k}|\mathbf{x})\log\Big(C_k(\mathbf{x})\Big)\bigg)d\mathbf{x}
\end{equation}

where $p(T_k|\mathbf{x})$ is the probability that $\mathbf{x}$ belongs to class $T_k$, which can be considered as the the $k$-th output of ``ground-truth'' classifier on sample $\mathbf{x}$ we expect the classifier $C$ to predict. Assume that $\Psi(C)$ has first-order derivative with respective to $C_k(\mathbf{x})$. The optimal solution of $C_k(\mathbf{x})$ can be obtained via setting this derivative equal to zero:

\begin{equation}
\begin{split}
\frac{\partial\Psi(C)}{\partial C_k(\mathbf{x})}
&= \frac{\partial}{\partial C_k(\mathbf{x})} \int {p_d^{T}}(\mathbf{x}) \bigg( \sum_{k=1}^{K}p({T_k}|\mathbf{x})\log\Big(C_k(\mathbf{x})\Big)\bigg)d\mathbf{x}\\
&= \frac{\partial}{\partial C_k(\mathbf{x})} \int {p_d^{T}}(\mathbf{x}) \bigg(p({T_1}|\mathbf{x})\log\Big(C_1(\mathbf{x})\Big) + \sum_{k=2}^{K}p({T_k}|\mathbf{x})\log\Big(C_k(\mathbf{x})\Big)\bigg)d\mathbf{x}\\
&= \frac{\partial}{\partial C_k(\mathbf{x})} \int {p_d^{T}}(\mathbf{x}) \bigg(p({T_1}|\mathbf{x})\log\Big(1 - \sum_{k=2}^{K}C_k(\mathbf{x})\Big) + \sum_{k=2}^{K}p({T_k}|\mathbf{x})\log\Big(C_k(\mathbf{x})\Big)\bigg)d\mathbf{x}\\
&= {p_d^T}(\mathbf{x}) \bigg( \frac{p({T_k}|\mathbf{x})}{C_k(\mathbf{x})} - \frac{p({T_1}|\mathbf{x})}{C_1(\mathbf{x})} \bigg)
\end{split}
\end{equation}

For any $k \in \{2, \dots, K\}$, setting $\frac{\partial\Psi}{\partial C_k(\mathbf{x})} = 0$, and the value of optimal $C_k^*$ has the following form:

\begin{equation}
\frac{p({T_1}|\mathbf{x})}{C^*_1(\mathbf{x})} = \frac{p({T_2}|\mathbf{x})}{C^*_2(\mathbf{x})} = \dots = \frac{p({T_k}|\mathbf{x})}{C^*_K(\mathbf{x})}
\label{ck_solution}
\end{equation}

Note that $\sum_{k=1}^{K}C^*_k(\mathbf{x}) = 1$, according to Bayes' theorem $p_d^T(\mathbf{x}) * p({T_k}|\mathbf{x}) = p(T_k) * p_d^{T_k}(\mathbf{x})$, and $p(T_i) = p(T_j) = \frac{1}{K}, i,j \in [1, K]$ (the probability we apply the transformations $T_k$ for sample $\mathbf{x}$ are equal), We finally obtain the optimal $C_k^*(\mathbf{x})$ from Eq. \ref{ck_solution}: $C^*_k(\mathbf{x}) = \frac{p({T_k}|\mathbf{x})}{\sum_{k=1}^{K}p({T_k}|\mathbf{x})} = \frac{p_d^{T_k}(\mathbf{x})}{\sum_{k=1}^{K}p_d^{T_k}(\mathbf{x})}$. That concludes our proof.

\noindent \textbf{Theorem \ref{theorem_1}} (Proof.) Substitute $C^*$ obtained above into $\Phi(G, C)$:

\begin{equation}
\Phi(G, C^*) = \int p_g^{T}(\mathbf{x})\bigg[\sum_{k=1}^{K}p({T_k}|\mathbf{x})\log\Big(C^*_k(\mathbf{x})\Big)\bigg]
\label{g_obj_1}
\end{equation}

Substitute  $C^*$ into (\ref{g_obj_1}) we have: 

\begin{equation}
\begin{split}
\Phi(G, C^*) &= \int p_g^{T}(\mathbf{x}) \sum_{k=1}^{K}\bigg[p({T_k}|\mathbf{x})\log\Big(C^*_k(\mathbf{x})\Big)\bigg]d\mathbf{x}\\
   &= \int p_g^{T}(\mathbf{x}) \sum_{k=1}^{K}\bigg[p({T_k}|\mathbf{x})\log\Big(\frac{p_d^{T_k}(\mathbf{x})}{\sum_{k=1}^{K}p_d^{T_k}(\mathbf{x})}\Big)d\mathbf{x}\bigg]\\
   &= \sum_{k=1}^{K}\int \bigg[p_g^{T}(\mathbf{x})p({T_k}|\mathbf{x})\log\Big(\frac{p_d^{T_k}(\mathbf{x})}{\sum_{k=1}^{K}p_d^{T_k}(\mathbf{x})}\Big)d\mathbf{x}\bigg]\\
   &= \sum_{k=1}^{K}\int \bigg[\frac{1}{K}p_g^{T_k}(\mathbf{x})\log\Big(\frac{p_d^{T_k}(\mathbf{x})}{\sum_{k=1}^{K}p_d^{T_k}(\mathbf{x})}\Big)d\mathbf{x}\bigg]\\
   &= \frac{1}{K}\sum_{k=1}^{K}\bigg[\int p_g^{T_k}(\mathbf{x})\log\Big(\frac{p_d^{T_k}(\mathbf{x})}{\sum_{k=1}^{K}p_d^{T_k}(\mathbf{x})}\Big)d\mathbf{x}\bigg]\\
   &= \frac{1}{K}\sum_{k=1}^{K}\bigg[\mathbb{E}_{\mathbf{x} \sim {P_g^{T_k}}}\log\Big(\frac{p_d^{T_k}(\mathbf{x})}{\sum_{k=1}^{K}p_d^{T_k}(\mathbf{x})}\Big)\bigg]
\end{split}
\label{g_obj_ss}
\end{equation}

That concludes our proof.
 

\noindent \textbf{Proposition \ref{prop_2}} (Proof.) Training self-supervised task $\Psi^+(G,C)$ with minimax game is similar to previous objective, except the additional term of fake class as below:

\begin{equation}
\begin{split}
\Psi^+(G,C) &= \mathbb{E}_{\mathbf{x} \sim P_d^T} \mathbb{E}_{T_k \sim \mathcal{T}}\log\Big(C_k(\mathbf{x})\Big) + \mathbb{E}_{\mathbf{x} \sim P_g^T}\mathbb{E}_{T_k \sim \mathcal{T}}\log\Big(C_{K+1}(\mathbf{x})\Big)\\
&= \int \Bigg(p_d^T(\mathbf{x})\sum_{i=1}^{K}p({T_k}|\mathbf{x})\log\Big(C_k(\mathbf{x})\Big) + p_g^T(\mathbf{x})\sum_{i=1}^{K}p({T_k}|\mathbf{x})\log\Big(C_{K+1}(\mathbf{x})\Big)\Bigg)d\mathbf{x}
\end{split}
\label{d_obj_ss_adv_1}
\end{equation}

Assume that $\Psi^+(G,C)$ has first-order derivative with respective to $C_k(\mathbf{x})$. The optimal $C^*_k(\mathbf{x})$ can be derived via setting derivative of $\Psi^+(G,C)$ equal to zero as follows:

\begin{equation}
\begin{split}
\frac{\partial\Psi^+(G,C)}{\partial C_k(\mathbf{x})} \\
= \frac{\partial}{\partial C_k(\mathbf{x})} &\int \Bigg(p_d^T(\mathbf{x})\sum_{i=1}^{K}p({T_k}|\mathbf{x})\log\Big(C_k(\mathbf{x})\Big) + p_g^T(\mathbf{x})\sum_{i=1}^{K}p({T_k}|\mathbf{x})\log\Big(C_{K+1}(\mathbf{x})\Big)\Bigg)d\mathbf{x}\\
= \frac{\partial}{\partial C_k(\mathbf{x})} &\int \bigg(p_d^T(\mathbf{x})p({T_1}|\mathbf{x})\log\Big(1 - \sum_{k=1}^{K}C_k(\mathbf{x}) - C_{K+1}(\mathbf{x})\Big)\\
&+ p_d^T(\mathbf{x})\sum_{k=2}^{K}p({T_k}|\mathbf{x})\log\Big(C_k(\mathbf{x})\Big) + {p_g^T(\mathbf{x})}\sum_{k=1}^{K}p({T_k}|\mathbf{x})\log\Big(C_{K+1}(\mathbf{x})\Big)\bigg)d\mathbf{x}
\end{split}
\end{equation}

Similar to above, for any $k \in \{2, \dots, K\}$, we have the derivative $\frac{\partial\Psi^+(G,C)}{\partial C_k(\mathbf{x})}$:

\begin{equation}
\frac{\partial\Psi^+(G,C)}{\partial C_k(\mathbf{x})} = p_d^T(\mathbf{x})\Big(\frac{p({T_1}|\mathbf{x})}{C^*_1(\mathbf{x})} - \frac{p({T_k}|\mathbf{x})}{C^*_k(\mathbf{x})}\Big)
\end{equation}

Setting $\frac{\partial\Psi}{\partial C_k(\mathbf{x})} = 0$, and we get optimal $C_k^*$, $k \in \{1, \dots, K\}$:

\begin{equation}
\frac{p_d^T(\mathbf{x})p({T_1}|\mathbf{x})}{C^*_1(\mathbf{x})} = \frac{p_d^T(\mathbf{x})p_{T_2}(\mathbf{x})}{C^*_2(\mathbf{x})} = \dots = \frac{p_d^T(\mathbf{x})p({T_k}|\mathbf{x})}{C^*_K(\mathbf{x})} = \frac{ p_d^T(\mathbf{x})\sum_{k=1}^{K}p({T_k}|\mathbf{x})}{\sum_{k=1}^{K}C^*_k(\mathbf{x})}
\label{ck_solution_adv}
\end{equation}

With $k = K+1$, we obtain the derivative of $\frac{\partial\Psi^+(G,C)}{\partial C_{K+1}(\mathbf{x})}$:

\begin{equation}
\frac{\partial\Psi^+(G,C)}{\partial C_{K+1}(\mathbf{x})} = p_d^T(\mathbf{x})\frac{p({T_1}|\mathbf{x})}{C^*_1(\mathbf{x})} - p_g^T(\mathbf{x})\frac{\sum_{k=1}{K}p({T_k}|\mathbf{x})}{C^*_{K+1}(\mathbf{x})}
\end{equation}

Setting $\frac{\partial\Psi}{\partial C_{K+1}(\mathbf{x})} = 0$, and finally we get optimal $C_k^*$, $k \in \{1, \dots, K+1\}$:

\begin{equation}
\begin{split}
\frac{p_d^T(\mathbf{x})p({T_1}|\mathbf{x})}{C^*_1(\mathbf{x})} &= \dots = \frac{p_d^T(\mathbf{x})p({T_k}|\mathbf{x})}{C^*_K(\mathbf{x})} = \frac{ p_d^T(\mathbf{x})\sum_{k=1}^{K}p({T_k}|\mathbf{x})}{\sum_{k=1}^{K}C^*_k(\mathbf{x})} = \frac{p_g^T(\mathbf{x})\sum_{k=1}^{K}p({T_k}|\mathbf{x})}{C^*_{K+1}(\mathbf{x})}
\end{split}
\label{ck_solution_adv_kpus1}
\end{equation}

Because $\sum_{k=1}^{K}C^*_k(\mathbf{x}) + C^*_{K+1}(\mathbf{x}) = 1$, we finally obtain the optimal $C_k^*(x)$ from Eq. \ref{ck_solution_adv}: $C^*_k(\mathbf{x}) = \frac{p_d^T(\mathbf{x})}{p_g^T(\mathbf{x})}\frac{p({T_k}|\mathbf{x})}{\sum_{k=1}^{K}p({T_k}|\mathbf{x})} C^*_{K+1}(\mathbf{x}) = \frac{p_d^T(\mathbf{x})}{p_g^T(\mathbf{x})}\frac{p_d^{T_k}(\mathbf{x})}{\sum_{k=1}^{K}p_d^{T_k}(\mathbf{x})} C^*_{K+1}(\mathbf{x})$. That concludes the proof.

\noindent \textbf{Theorem \ref{theorem_2}} (Proof.) Substitute optimal $C^*$ obtained above into  $\Phi^+(G,C)$:

\begin{equation}
\Phi^+(G,C^*) = \bigg(\mathbb{E}_{\mathbf{x} \sim {P_g^T}}\sum_{k=1}^{K}p({T_k}|\mathbf{x})\log\Big(C^*_k(\mathbf{x})\Big) - \mathbb{E}_{\mathbf{x} \sim P_g^T}\sum_{k=1}^{K}p({T_k}|\mathbf{x})\log\Big(C^*_{K+1}(\mathbf{x})\Big)\bigg)
\end{equation}

The first term can be written as:

\begin{equation}
\begin{split}
&\mathbb{E}_{\mathbf{x} \sim {P_g^T}}\sum_{k=1}^{K}p({T_k}|\mathbf{x})\log(C^*_k(\mathbf{x}))\\
&= \mathbb{E}_{\mathbf{x} \sim {P_g^T}}\bigg[\sum_{k=1}^{K}p({T_k}|\mathbf{x})\log\Big(\frac{p_d^T(\mathbf{x})}{p_g^T(\mathbf{x})}\frac{p({T_k}|\mathbf{x})}{\sum_{k=1}^{K}p({T_k}|\mathbf{x})} C^*_{K+1}(\mathbf{x})\Big)\bigg] \\
&= \mathbb{E}_{\mathbf{x} \sim P_g^T}\bigg[\sum_{k=1}^{K}p({T_k}|\mathbf{x})\log\Big(C^*_{K+1}(\mathbf{x})\Big) + \log\Big(\frac{p_d^T(\mathbf{x})}{p_g^T(\mathbf{x})}\Big) + \log\Big(\frac{p({T_k}|\mathbf{x})}{\sum_{k=1}^{K}p({T_k}|\mathbf{x})}\Big)\bigg]\\
&= \mathbb{E}_{\mathbf{x} \sim P_g^T}\bigg[\sum_{k=1}^{K}p({T_k}|\mathbf{x})\log\Big(C^*_{K+1}(\mathbf{x})\Big)\bigg] + \mathbb{E}_{\mathbf{x} \sim P_g^T}\bigg[\sum_{k=1}^{K}p({T_k}|\mathbf{x})\log\Big(\frac{p_d^T(\mathbf{x})}{p_g^T(\mathbf{x})}\Big)\bigg] \\
&\quad \quad \quad \quad \quad \quad \quad \quad \quad \quad \quad \quad + \mathbb{E}_{\mathbf{x} \sim P_g^T}\bigg[\sum_{k=1}^{K}p({T_k}|\mathbf{x})\log\Big(\frac{p({T_k}|\mathbf{x})}{\sum_{k=1}^{K}p({T_k}|\mathbf{x})}\Big)\bigg]\\
&= \mathbb{E}_{\mathbf{x} \sim P_g^T}\bigg[\sum_{k=1}^{K}p({T_k}|\mathbf{x})\log\Big(C^*_{K+1}(\mathbf{x})\Big)\bigg] +  \frac{1}{K}\bigg[\sum_{k=1}^{K}\int p_g^{T_k}(\mathbf{x})\log\Big(\frac{p_d^{T_k}(\mathbf{x})}{p_g^{T_k}(\mathbf{x})}\Big)d\mathbf{x}\bigg] \\
&\quad \quad \quad \quad \quad \quad \quad \quad \quad \quad \quad \quad + \mathbb{E}_{\mathbf{x} \sim P_g^T}\bigg[\sum_{k=1}^{K}p({T_k}|\mathbf{x})\log\Big(\frac{p({T_k}|\mathbf{x})*p_d^T(\mathbf{x})}{\sum_{k=1}^{K}p({T_k}|\mathbf{x})*p_d^T(\mathbf{x})}\Big)\bigg]\\
&= \mathbb{E}_{\mathbf{x} \sim P_g^T}\bigg[\sum_{k=1}^{K}p({T_k}|\mathbf{x})\log\Big(C^*_{K+1}(\mathbf{x})\Big)\bigg] - \frac{1}{K}\bigg[\sum_{k=1}^{K}\mathrm{KL}(P_g^{T_k}||P_d^{T_k})\bigg] \\
&\quad \quad \quad \quad \quad \quad \quad \quad \quad \quad \quad \quad + \frac{1}{K}\sum_{k=1}^{K}\bigg[\mathbb{E}_{\mathbf{x} \sim P_g^{T_k}}\log\Big(\frac{p_d^{T_k}(\mathbf{x})}{\sum_{k=1}^{K}p_d^{T_k}(\mathbf{x})}\Big)\bigg]\\
\end{split} 
\label{g_obj_lower_bound}
\end{equation}

With the note that $p_d^T(\mathbf{x}) * p({T_k}|\mathbf{x}) = p(T_k) * p_d^{T_k}(\mathbf{x}) = \frac{1}{K}p_d^{T_k}(\mathbf{x})$ and $p_g^T(\mathbf{x}) * p({T_k}|\mathbf{x}) = p(T_k) * p_g^{T_k}(\mathbf{x}) = \frac{1}{K}p_g^{T_k}(\mathbf{x})$. Moving the first term of Eq. \ref{g_obj_lower_bound} from the right side to left side, it concludes the proof.

\begin{theorem}
KL divergence is invariant to affine transform.
\end{theorem}

\textit{Proofs}. Let $\mathbf{x}\in\mathbb{R}^{n\times 1}$ be a random variable. $p_x(\mathbf{x})$
is a distribution defined on $\mathbf{x}$. Let $T$ be an affine
transform, i.e., $T(\mathbf{x})=\boldsymbol{A}\mathbf{x}+\boldsymbol{b}$,
where $\boldsymbol{A}\in\mathbb{R}^{n\times n}$ is a full rank matrix and $\boldsymbol{b}\in\mathbb{R}^{n\times 1}$.
Then for a random variable $\mathbf{y}=T(\mathbf{x})=\boldsymbol{A}\mathbf{x}+\boldsymbol{b}$,
$p_y(\mathbf{y})=|\boldsymbol{J}|p_x\Big(T^{-1}(\mathbf{y})\Big)$, where $\boldsymbol{J}$
is the Jacobian matrix, with its $(i,j)$-th entry defined as:
\begin{align}
\boldsymbol{J}_{i,j}=\frac{\partial \mathbf{x}_i}{\partial \mathbf{y}_j}
\end{align}
Obviously, $\boldsymbol{J}=\boldsymbol{A}^{-1}$. Then we have
$p_y(\mathbf{y})=|\boldsymbol{A}^{-1}|p_x\Big(T^{-1}(\mathbf{y})\Big)$.

Let $p_{x_1}(\mathbf{x})$ and $p_{x_2}(\mathbf{x})$ are two distributions
defined on $\mathbf{x}$. Then let $p_{y_1}(\mathbf{y})$ and $p_{y_2}(\mathbf{y})$
be the corresponding distributions defined on $\mathbf{y}$.
Then we have $p_{y_1}(\mathbf{y})=|\boldsymbol{A}^{-1}|p_{x_1}\Big(T^{-1}(\mathbf{y})\Big)$
and $p_{y_2}(\mathbf{y})=|\boldsymbol{A}^{-1}|p_{x_2}\Big(T^{-1}(\mathbf{y})\Big)$.

Using the definition of the KL divergence between $p_{y_1}$ and $p_{y_2}$,
we have:

\begin{align}
\mathrm{KL}(p_{y_1}||p_{y_2})&=\int p_{y_1}(\mathbf{y})\log\frac{p_{y_1}(\mathbf{y})}{p_{y_2}(\mathbf{y})}d\mathbf{y}\\
&=\int|\boldsymbol{A}^{-1}| p_{x_1}\Big(T^{-1}(\mathbf{y})\Big)\log\frac{|\boldsymbol{A}^{-1}|p_{x_1}(T^{-1}(\mathbf{y}))}{|\boldsymbol{A}^{-1}|p_{x_2}\Big(T^{-1}(\mathbf{y})\Big)}d\mathbf{y}\\
&=\int|\boldsymbol{A}^{-1}| p_{x_1}\Big(T^{-1}(\mathbf{y})\Big)\log\frac{p_{x_1}\Big(T^{-1}(\mathbf{y})\Big)}{p_{x_2}\Big(T^{-1}(\mathbf{y})\Big)}d\mathbf{y}
\end{align}

As $\mathbf{x}=T^{-1}(\mathbf{y})$, then we have:

\begin{align}
\mathrm{KL}(p_{y_1}||p_{y_2})
&=\int|\boldsymbol{A}^{-1}| p_{x_1}\Big(T^{-1}(\mathbf{y})\Big)\log\frac{p_{x_1}\Big(T^{-1}(\mathbf{y})\Big)}{p_{x_2}\Big(T^{-1}(\mathbf{y})\Big)}d\mathbf{y}\\
&=\int|\boldsymbol{A}^{-1}| p_{x_1}(\mathbf{x})\log\frac{p_{x_1}(\mathbf{x})}{p_{x_2}(\mathbf{x})}d\mathbf{y}
\end{align}

According to the property of multiple integral, we have:

\begin{align}
\mathrm{KL}(p_{y_1}||p_{y_2})
&=\int|\boldsymbol{A}^{-1}| p_{x_1}(\mathbf{x})\log\frac{p_{x_1}(\mathbf{x})}{p_{x_2}(\mathbf{x})}|\boldsymbol{A}|d\mathbf{x}\\
&=\int p_{x_1}(\mathbf{x})\log\frac{p_{x_1}(\mathbf{x})}{p_{x_2}(\mathbf{x})}d\mathbf{x}\\
&=\mathrm{KL}(p_{x_1}||p_{x_2})
\end{align}

It concludes our proof.

\begin{corollary}
KL divergence between real and fake distributions is equal to that of rotated real and rotated fake distributions by $T_k$: $\mathrm{KL}(P_g^{T_k}||P_d^{T_k}) = \mathrm{KL}(P_g||P_d), k \in [1 : K]$
\end{corollary}

Note that we apply the above theorem of invariance of KL, with $p_{x_1},p_{x_2}$ being $P_g, P_d$ respectively, and image rotation $T_k$ as the transform.

\subsection{Implementation}

Here, we discuss  details of our implementation. For the SS tasks, we follow the geometric transformation of \cite{gidaris-iclr-2018} to argument images and compute pseudo labels. It is simple yet effective and currently the state-of-the-art in self-supervised tasks. In particular, we train discriminator to recognize the 2D rotations which were applied to the input image. We rotate the input image with $K = 4$ rotations ($0^{\circ}, 90^{\circ}, 180^{\circ}, 270^{\circ}$) and assign them the pseudo-labels from 1 to $K$.

To implement our model, the GAN objectives for discriminator and generator can be the ones in original  GAN by Goodfellow et al. \cite{goodfellow-nisp-2014}, or other variants. In our work, we conduct experiments to show improvements with two baseline models: original SSGAN \cite{chen-arxiv-2018} and DistGAN \cite{tran-eccv-2018}.

We integrate SS tasks into Dist-GAN \cite{tran-eccv-2018} and conduct  study with this baseline. 
In our experiments, we observe that Dist-GAN has good convergence property and  this is important for our ablation study.



\begin{equation}
\begin{split}
\min_G\mathcal{V}(D,C,G) &= \mathcal{V}(D,G)\\
+ \lambda_g \bigg|\bigg|&\bigg(\mathbb{E}_{\mathrm{x} \sim {P_g^T}}\mathbb{E}_{T_k \sim \mathcal{T}}\log\Big(C_k(\mathrm{x})\Big) - \mathbb{E}_{\mathrm{x} \sim P_g^T}\mathbb{E}_{T_k \sim \mathcal{T}}\log\Big(C_{K+1}(\mathrm{x})\Big)\bigg) \\
&- \bigg(\mathbb{E}_{\mathrm{x} \sim {P_d^T}}\mathbb{E}_{T_k \sim \mathcal{T}}\log\Big(C_k(\mathrm{x})\Big) - \mathbb{E}_{\mathrm{x} \sim P_d^T}\mathbb{E}_{T_k \sim \mathcal{T}}\log\Big(C_{K+1}(\mathrm{x})\Big)\bigg)\bigg|\bigg|
\end{split}
\label{g_ss_maximum_entropy_loss}
\end{equation}

Second, in practice, achieving equilibrium point for optimal D, G, C is difficult. Therefore, inspired by \cite{tran-eccv-2018}, we propose the new generator objective to improve Eq. \ref{gan_gen_obj_ss_adv} as written in Eq. \ref{g_ss_maximum_entropy_loss}. It couples the convergence of $\Phi^+(G,C)$ and $\Psi^+(G,C)$ that allows the learning is more stable. Our intuition is that if generator distribution is similar to the real distribution, the classification performance on its transformed fake samples should be similar to that of those from real samples. Therefore, we propose to match the self-supervised tasks of real and fake samples to train the generator. In other words, if real and fake samples are from similar distributions, the same tasks applied for real and fake samples should have resulted in similar behaviors. In particular, given the cross-entropy loss computed on real samples, we train the generator to create samples that are able to match this loss. Here, we use $\ell_1$-norm for the $\Phi^+(G,C)$ and $\mathcal{V}(D,G)$ is the objective of GAN task \cite{tran-eccv-2018}. In our implementation, we randomly select a geometric transformation $T_k$ for each data sample when training the discriminator. And the same $T_k$ are applied for generated samples when matching the self-supervised tasks to train the generator. 

%

For this objective of generator, similar to Eq. \ref{g_obj_lower_bound}, we have: 

\begin{equation}
\begin{split}
\mathbb{E}_{\mathbf{x}\sim {P_d^T}}\bigg[\sum_{k=1}^{K}p({T_k}|\mathbf{x})&\log\Big(C^*_k(\mathbf{x})\Big)\bigg]\\
&= \mathbb{E}_{\mathbf{x} \sim P_d^T}\bigg[\sum_{k=1}^{K}p({T_k}|\mathbf{x})\log\Big(C^*_{K+1}(\mathbf{x})\Big)\bigg]+ \bigg[\sum_{k=1}^{K}\mathrm{KL}(P_d^{T_k}||P_g^{T_k})\bigg]\\
&+ \sum_{k=1}^{K}\bigg[\mathbb{E}_{\mathbf{x} \sim P_d^{T_k}}\log\Big(\frac{p_d^{T_k}(\mathbf{x})}{\sum_{k=1}^{K}p_d^{T_k}(\mathbf{x})}\Big)\bigg]\\
\end{split} 
\label{d_obj_lower_bound}
\end{equation}

The objective of Eq. \ref{g_ss_maximum_entropy_loss} can be re-written as:

\begin{equation}
\begin{split}
* 
&= \bigg|\bigg|\sum_{k=1}^{K}\Big(\mathrm{KL}(P_d^{T_k}||P_g^{T_k}) + \mathrm{KL}(P_g^{T_k}||P_d^{T_k})\Big) \\
&+ \sum_{k=1}^{K}\mathbb{E}_{\mathbf{x} \sim P_g^{T_k}}\log\Big(\frac{p_d^{T_k}(\mathbf{x})}{\sum_{k=1}^{K}p_d^{T_k}(\mathbf{x})}\Big)- \sum_{k=1}^{K}\mathbb{E}_{\mathbf{x} \sim P_d^{T_k}}\log\Big(\frac{p_d^{T_k}(\mathbf{x})}{\sum_{k=1}^{K}p_d^{T_k}(\mathbf{x})}\Big)\bigg|\bigg| \geq 0
\end{split}
\label{g_obj_bound}
\end{equation}

$P_g = P_d$ is the solution that minimizes Eq. \ref{g_obj_bound}. 
In practice, we found that this is stable. It is due to the stability of symmetric KL divergence (forward KL and inverse KL).

\section{Appendix: Experiments}
\label{appendix_b}


\begin{figure}
  \centering
  \includegraphics[scale=0.74]{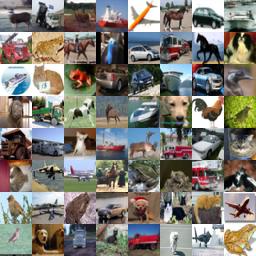}
  \includegraphics[scale=0.74]{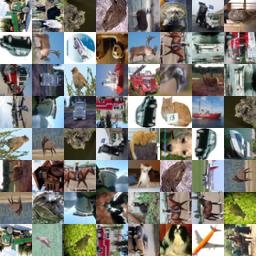}
  \includegraphics[scale=0.74]{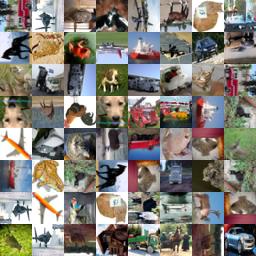}
  \includegraphics[scale=0.74]{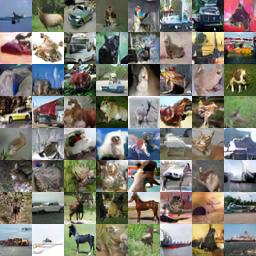}
  \caption{From left to right: Real samples, argument real samples by rotation, mixed argument real and fake samples, and generated images of CIFAR-10.}
  \label{cifar_resnet_samples}
\end{figure}

\begin{figure}
  \centering
  \includegraphics[scale=0.50]{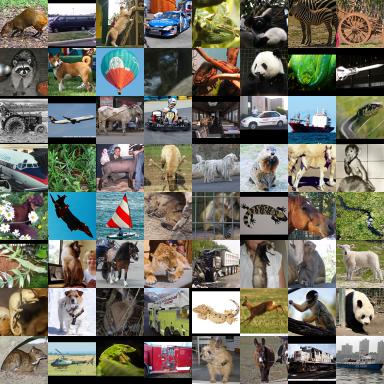}
  \includegraphics[scale=0.50]{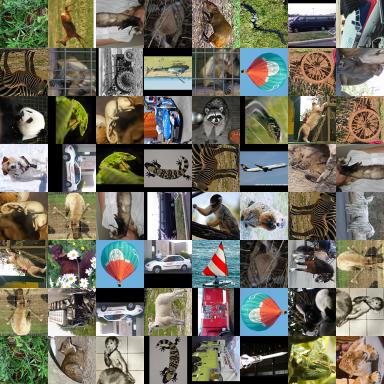}
  \includegraphics[scale=0.50]{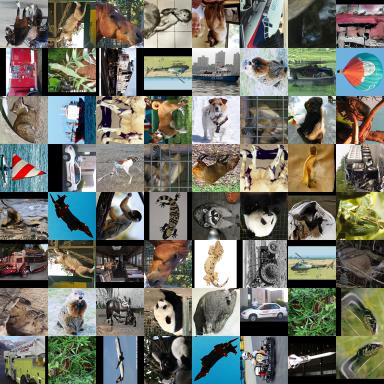}
  \includegraphics[scale=0.50]{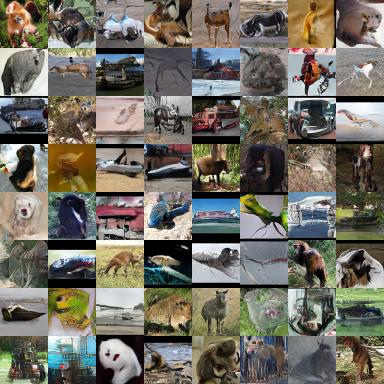}
  \caption{From left to right: Real samples, argument real samples by rotation, mixed argument real and fake samples, and generated images of STL-10.}
  \label{stl_resnet_samples}
\end{figure}

\subsection{Details of experiment setup}
\label{ablation_study_ld}

In our experiments, FID is computed every 10K iterations in training and visualized with the smoothening windows of 5. The latent dimension is $d_\mathrm{z} = 128$ and mini-batch size is 64 for our all experiments. We visualize losses and FID scores in several figures. In these figures, the horizontal axis is the number of training iterations, and the vertical axis is either the loss and FID score. We compute the negative  discriminator/classifier value function for the visualization. We investigate the improvements of our proposed techniques on two baseline models:

Dist-GAN \cite{tran-eccv-2018}: We use Dist-GAN implemented with three network architectures: DCGAN, CNN in SN-GAN and ResNet.
We use standard ``log" loss for DCGAN architecture, and  with ``hinge" loss SN-GAN (the CNN network as in SN-GAN \cite{miyato-iclr-2018}) and ResNet architectures. We use ``hinge" loss for SN-GAN  and ResNet because it attains better performance than standard ``log" loss as shown in \cite{miyato-iclr-2018}. 
We train models using Adam optimizer with learning rate $\mathrm{lr} = 0.0002$, $\beta_1 = 0.5$, $\beta_2 = 0.9$ for DCGAN and SN-GAN architectures and $\beta_1 = 0.0$, $\beta_2 = 0.9$ for ResNet architecture \cite{gulrajani-arxiv-2017}. {\bf If not precisely mentioned, it means Dist-GAN is used for the experiments.}

SSGAN: We were unable to reproduce results as reported in the original paper with this code\footnote{https://github.com/google/compare\_gan}, although we have followed the best parameter settings of the paper and communicated with authors of SSGAN regarding the issues. We achieve best results with another setting (spectral norm, $\lambda = 0, \mathrm{d_{iter}} = 10, \beta_1 = 0.5, \beta_2 = 0.999$).
We use this setting as the baseline and compare to the one using our proposed SS tasks instead of the original SS tasks.

\subsection{Ablation study SS in Discriminator and Generator Learning for the original SS proposed in 
\cite{chen-arxiv-2018}}

\label{appendix_ablation_discriminator_learning}

In this experiment, we analyze original SS tasks proposed in \cite{chen-arxiv-2018} to understand the effect of self-supervised tasks.
We aim to provide empirical observation of how the $\Psi(C)$ contributes to the discriminator via changing $\lambda_d$ with fixed $\lambda_g = 0$. Experiments are on CIFAR-10 dataset using small DCGAN architecture. For implementation, they are integrated into the discriminator of the baseline model, Dist-GAN \cite{tran-eccv-2018} as mentioned above. Through the experiment, we confirm that the contribution of $\Psi(C)$ is important in Dist-GAN model. We should set the $\lambda_d$ attain the good trade-off between GAN task and SS task because increasing $\lambda_d$ is not helpful. The SS task with $\lambda_d = 1.0$ is good for Dist-GAN model, which is also discussed in \cite{chen-arxiv-2018} with SN-GAN model \cite{miyato-iclr-2018}. 


\begin{figure}
  \centering
  \includegraphics[width=6.5cm,keepaspectratio]{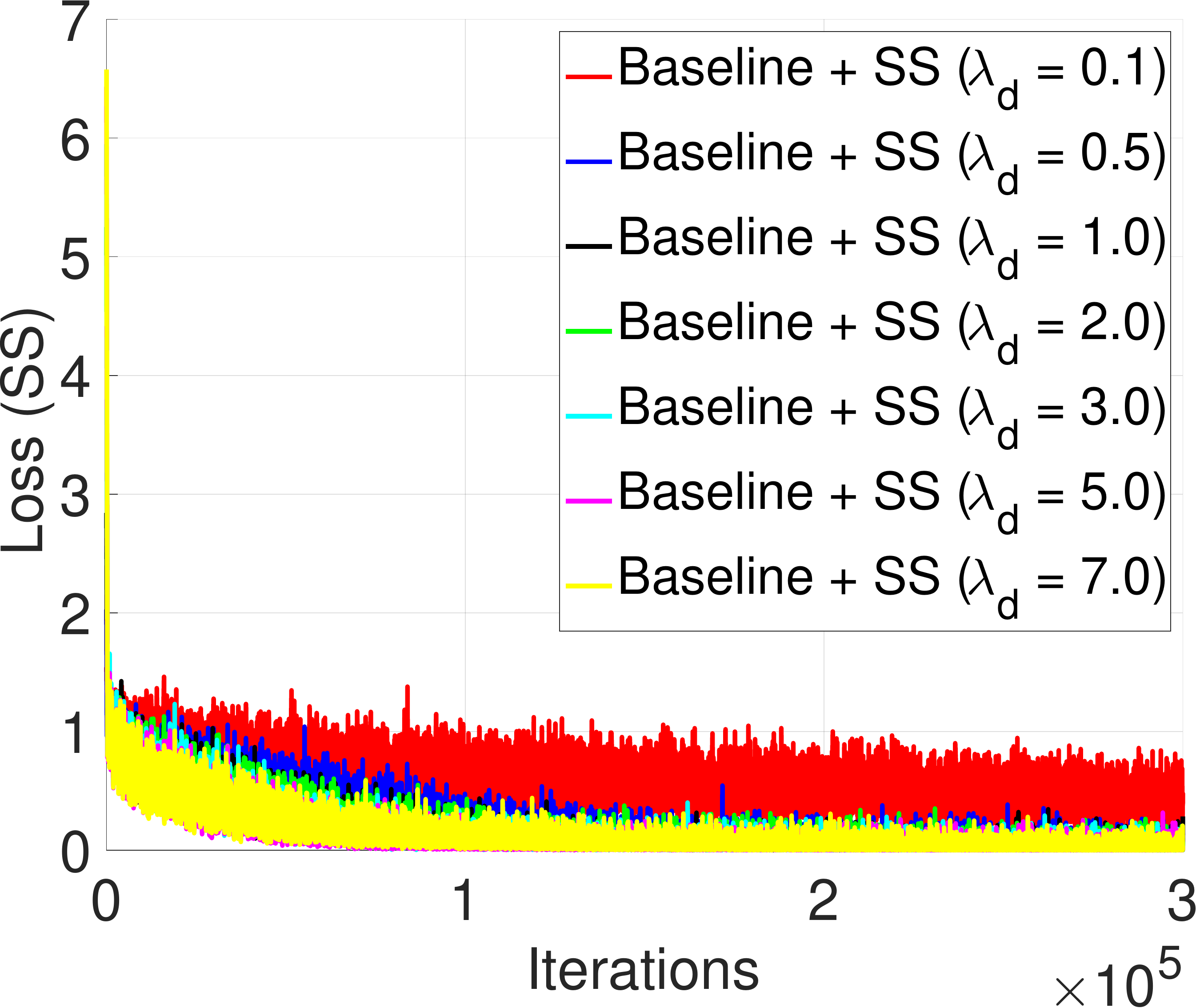}
  \includegraphics[width=6.5cm,keepaspectratio]{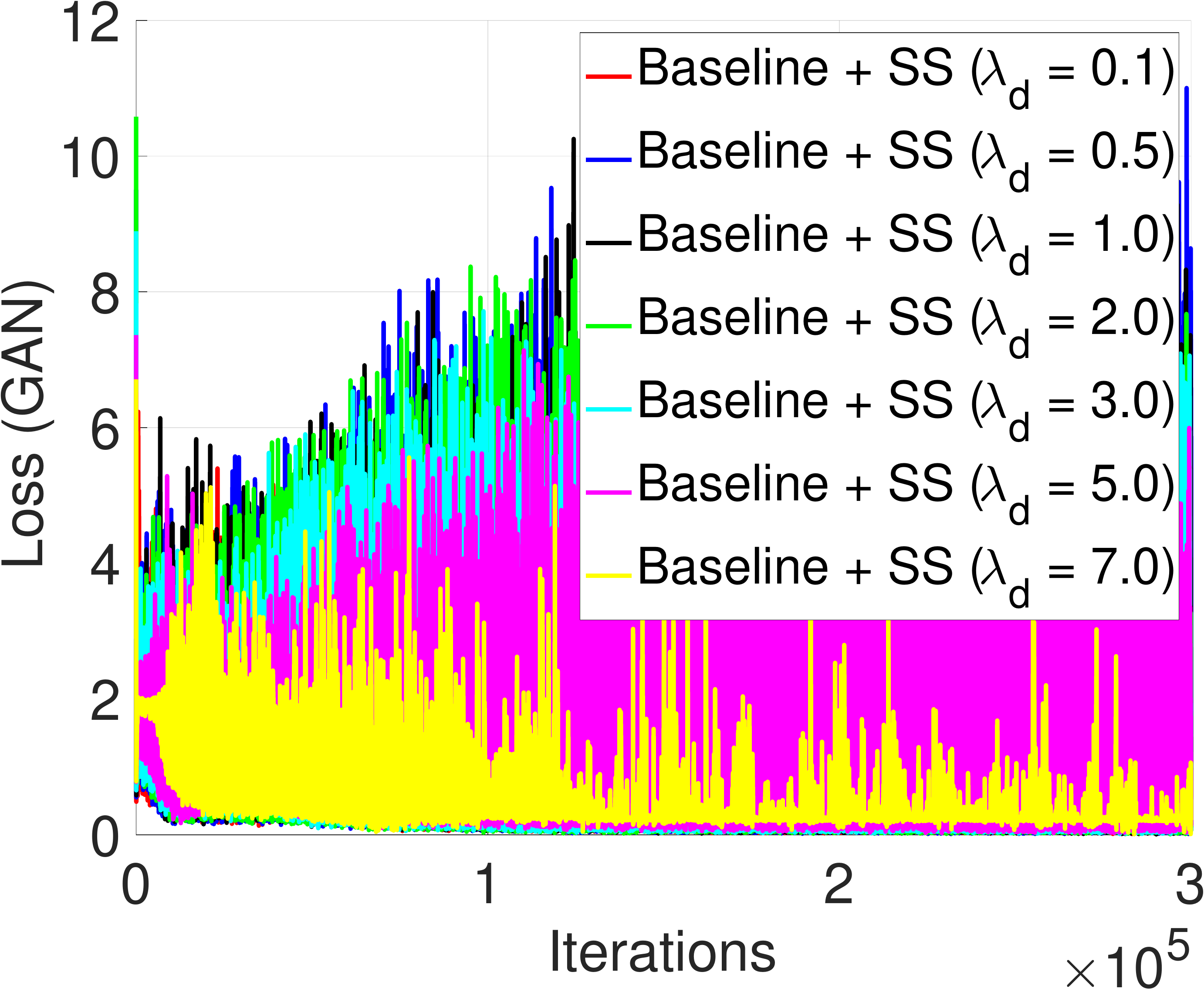}
  \includegraphics[width=6.5cm,keepaspectratio]{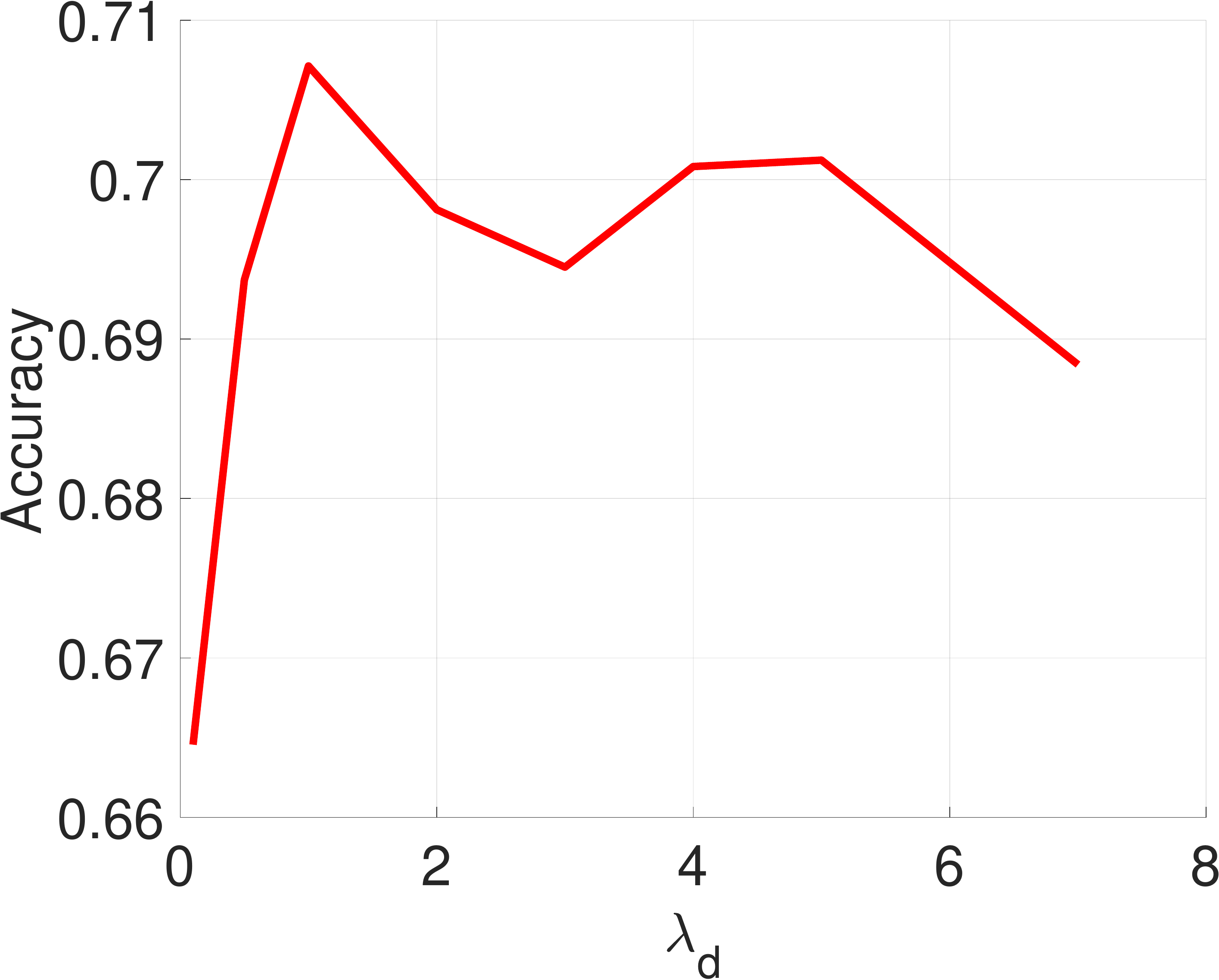}
  \includegraphics[width=6.5cm,keepaspectratio]{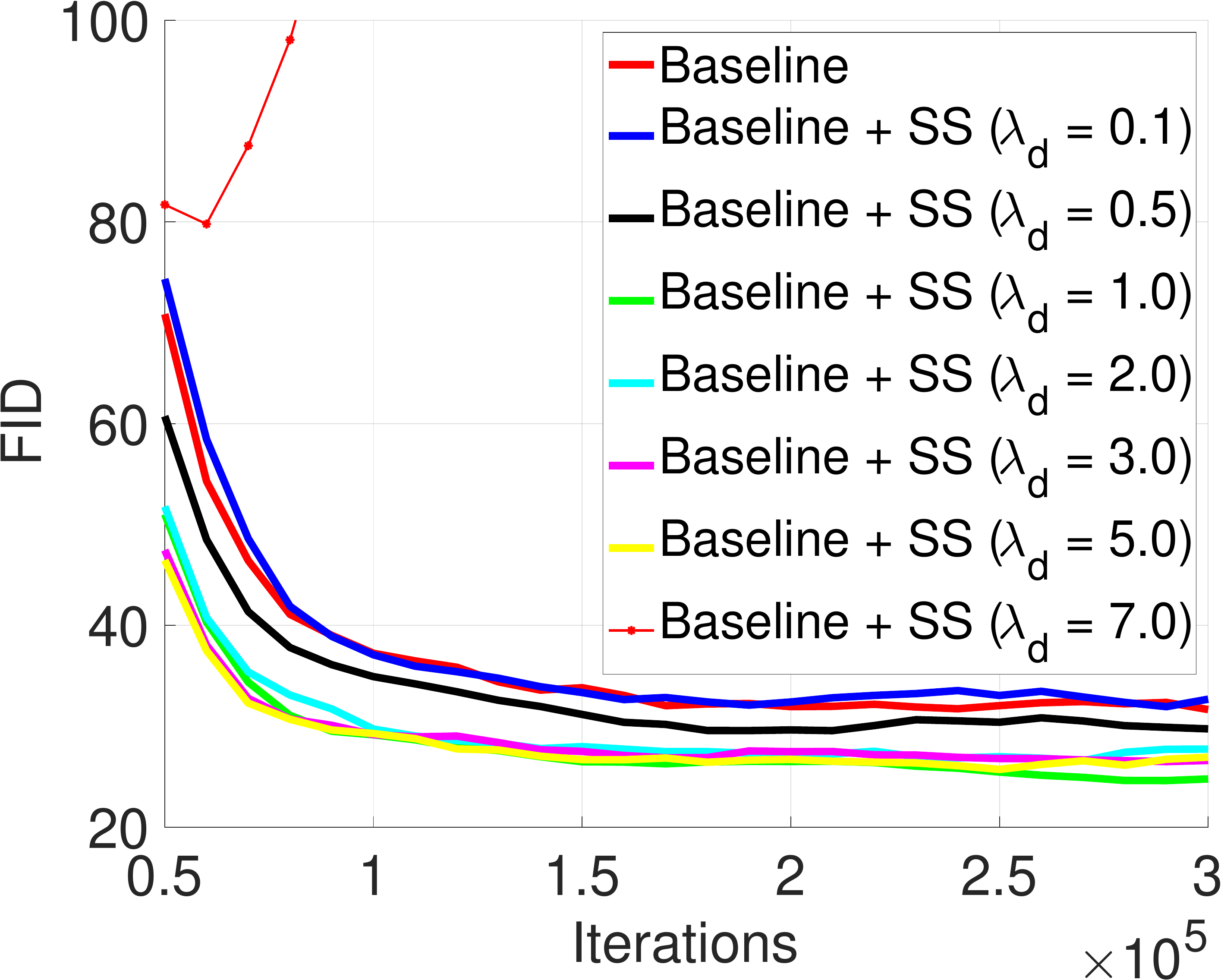}
  \caption{The ablation study of SS task $\Psi(C)$ as proposed in \cite{chen-arxiv-2018}. We analyze its effect via $\lambda_d$ fine-tuning, $\lambda_g = 0.0$. (a) The discriminator losses of SS task, (b) The discriminator losses of GAN task, (c) the feature representation quality and (d) FID scores. With $\lambda_d = 7.0$ for SS task, the model becomes seriously collapsed with FID > 100. Experiments are conducted with the baseline model, Dist-GAN. (Best view in color).}
  \label{ss_finetuning_dcgan}
\end{figure}
 
The results in Fig. \ref{ss_finetuning_dcgan} illustrate the effects of $\Psi(C)$ to GAN with different values of $\lambda_d$. Fig. \ref{ss_finetuning_dcgan}a represents the losses of the SS task of the discriminator. It shows that in most cases, the larger $\lambda_d$ lead to faster $\Psi(C)$ loss converges. However, when $\lambda_d > 1.0$, the FID is not improved. We observe that once $\lambda_d$ is higher, the loss of GAN task is dominated by the SS tasks. When $\lambda_d$ is too high, (e.g., $\lambda_d = 7.0$), GAN loss is almost unchanged about first 10K iterations (Fig. \ref{ss_finetuning_dcgan}b) in early iterations and the model gets collapsed. 
This can be explained as follows.
When the discriminator improvement is slow due to the strong dominance of $\Psi(C)$, the learning of the generator faster. This serious unbalance easily leads to the collapsed generator and the learning of generator gets stuck thereafter. When the GAN loss is strongly dominated by SS loss, the loss of GAN is saturated.

To understand deeper, we evaluate the representation qualities of the intermediate layers of the discriminator as in  \cite{chen-arxiv-2018} in this experiment. Given the above pre-trained discriminators, we compute features of train and test sets of CIFAR-10 via its last convolution layer. We evaluate the classification performance as training logistic regression on these features and measure with top-1 accuracy. We follow the experimental setup of parameters as in \cite{chen-arxiv-2018}. The result (Fig. \ref{ss_finetuning_dcgan}c) that as $\lambda_d$ >= 1.0, the accuracy is also similar, except for the case $\lambda = 7.0$, the quality of feature is slightly worse but not too significant (although the GAN model is collapsed). It means increasing $\lambda_d$ does not necessarily improve the feature representation quality of the discriminator. 

Overall, 
with Dist-GAN as baseline, we observe that using the original SS tasks with $\lambda_d = 1.0$ provide considerable improvement, and the results suggest that $\lambda_d$ should not be too high, but instead the one that provides a good trade-off between GAN and SS tasks. 

\subsubsection{SS task in Generator Learning}
\label{ss_task_generator_learning}

\begin{figure}
  \centering
  \includegraphics[width=4.5cm,keepaspectratio]{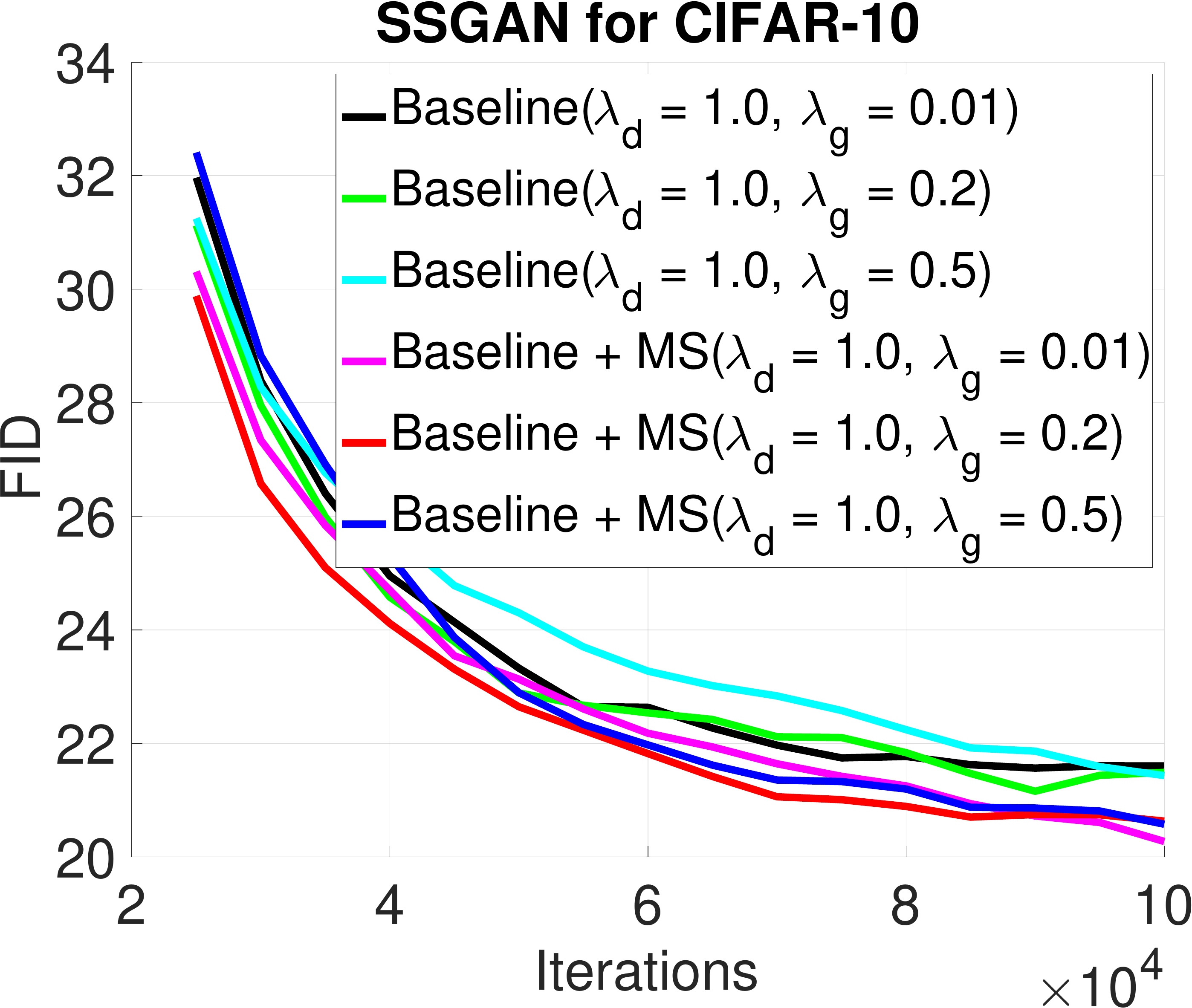}
  \includegraphics[width=4.5cm,keepaspectratio]{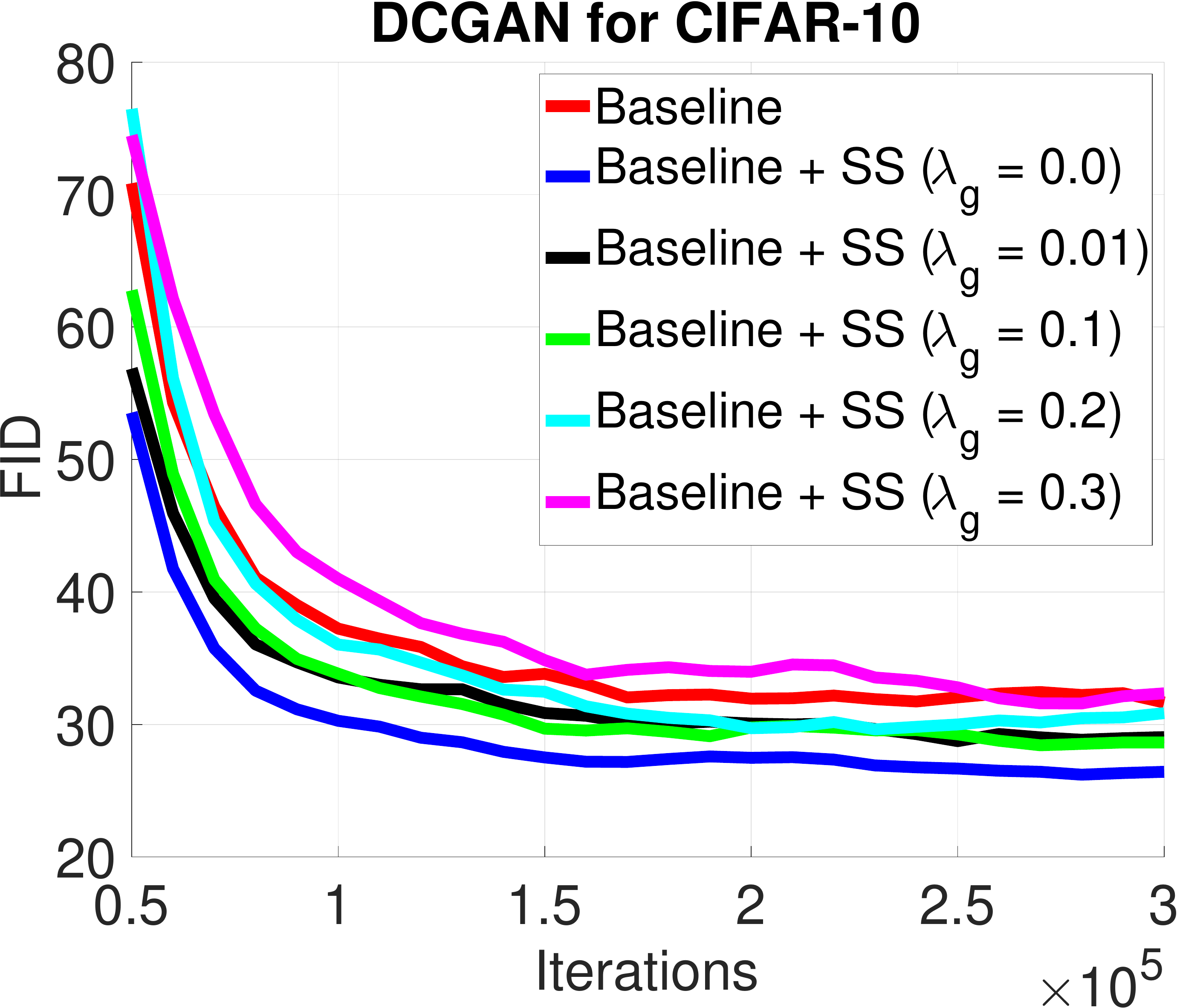}
  \includegraphics[width=4.5cm,keepaspectratio]{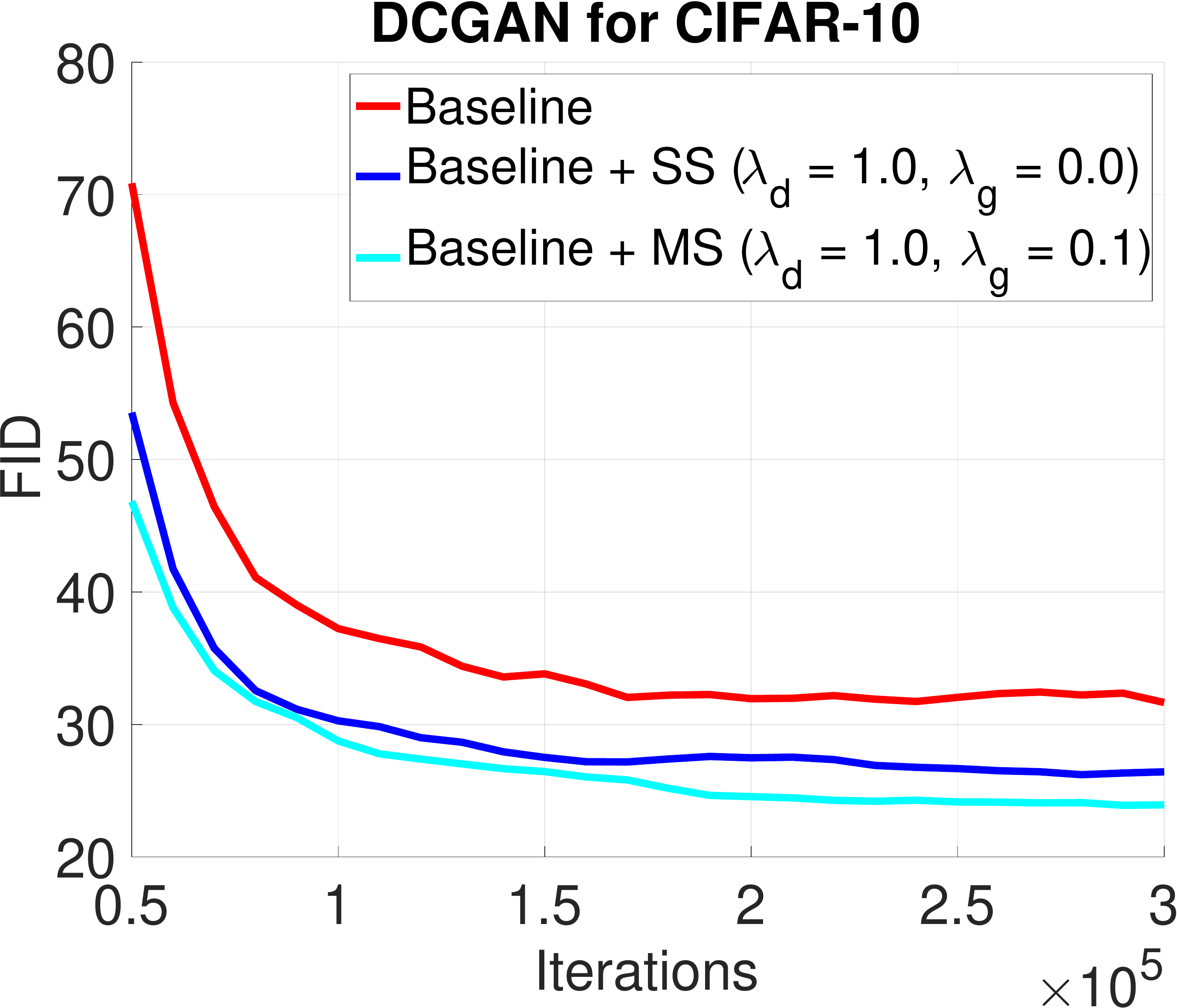}
  \caption{The ablation studies with (a) SSGAN$^+$ using $\lambda_g$ and ours (SSGAN$^+$ + MS) (b) Dist-GAN with $\Phi(C)$ and $\Psi(G,C)$ as fine-tuning $\lambda_g$, fixed $\lambda_d = 1.0$. (c) Our model (Dist-GAN + MS) with $\Psi^+(G,C)$ and $\Phi^+(G,C)$ with $\lambda_g = 0.1$ and $\lambda_d = 1.0$. Experiments are with CIFAR-10 dataset. (Best view in color)}
  \label{cifar_dcgan_dinetuning_g_d}
\end{figure}

We continue to investigate the effects of $\lambda_g$ with fixed $\lambda_d = 1.0$
for the SS tasks proposed in 
\cite{chen-arxiv-2018}.
The experimental setup is similar to the previous one. The result represented in Fig. \ref{cifar_dcgan_dinetuning_g_d}a show that $\lambda_g > 0$ still improves the baseline model, but higher than the case of $\lambda_g = 0$. 
Note that 
\cite{chen-arxiv-2018} does not report result with $\lambda_g = 0.0$.


Following our discussion 
on  Theorem \ref{theorem_1}, applying $\Phi(G,C)$ as proposed 
in 
\cite{chen-arxiv-2018}
does not support the matching between the generator and data distributions. From these experiments, we observe that the generator and discriminator are unable to reach optimal points, and using large $\lambda_g$ degrades the quality of GAN task, and even leads to mode collapse. For example, as  $\lambda_g$ increases (eg. $\lambda_g = 0.3$), it seriously hurts the quality of GAN task of the generator.

In addition, we verify with original code of SSGAN \cite{chen-arxiv-2018} on CIFAR-10 using our best setting mentioned above. Fig. \ref{cifar_dcgan_dinetuning_g_d}a confirms that with our best setting $\lambda_g = 0.01$ and $\lambda_g = 0.2$ achieve similar FID and increasing $\lambda_g = 0.5$ degrades its performance, which is consistent to our analysis. In the same figure, when we use our proposed MS,  FID is improved.



\subsection{Ablation study ($\lambda_d$, $\lambda_g$) with DCGAN for our proposed method}
\label{ablation_study_minimax_dcgan}

\begin{figure}
  \centering
  \includegraphics[width=6.5cm,keepaspectratio]{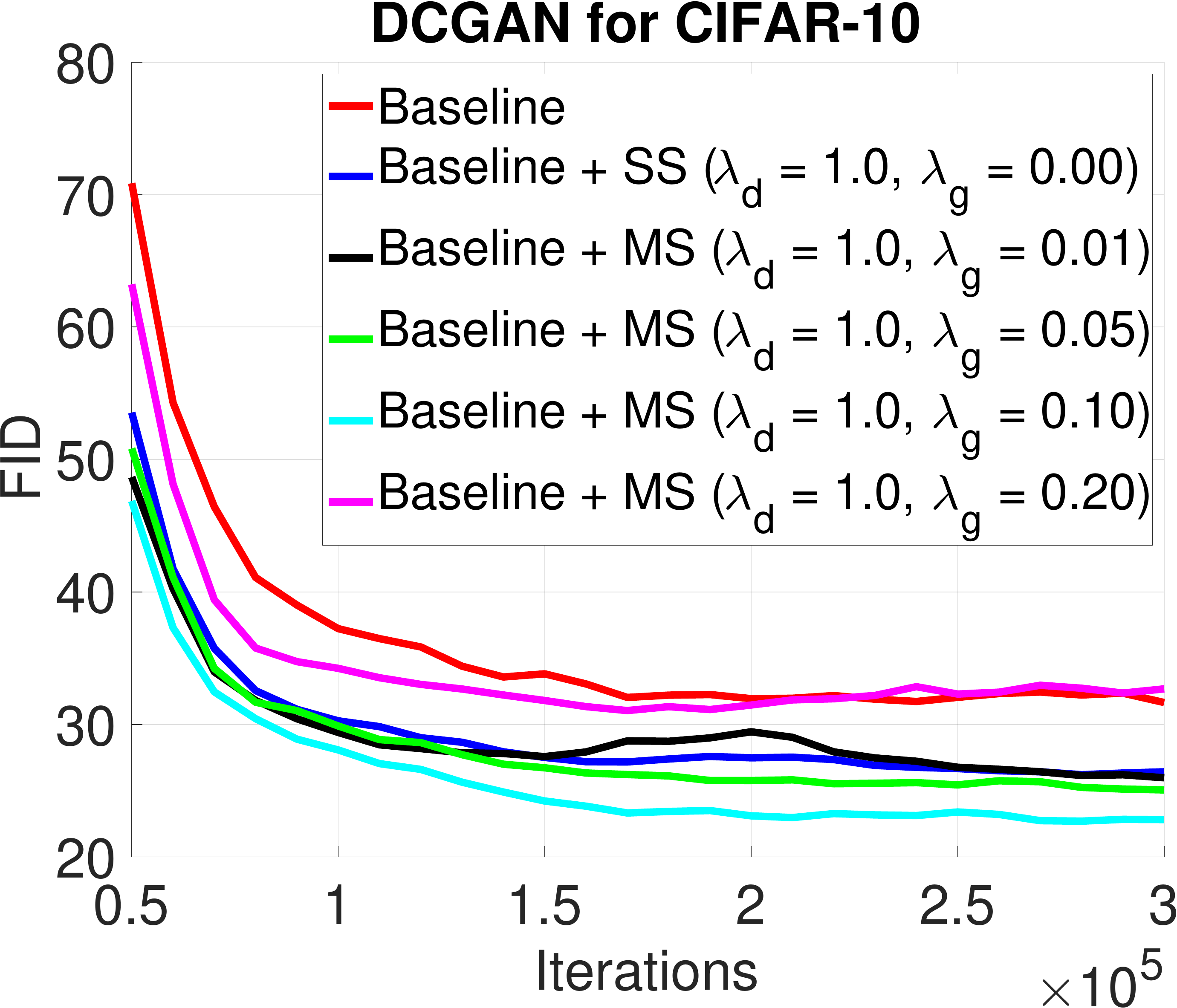}
  \includegraphics[width=6.5cm,keepaspectratio]{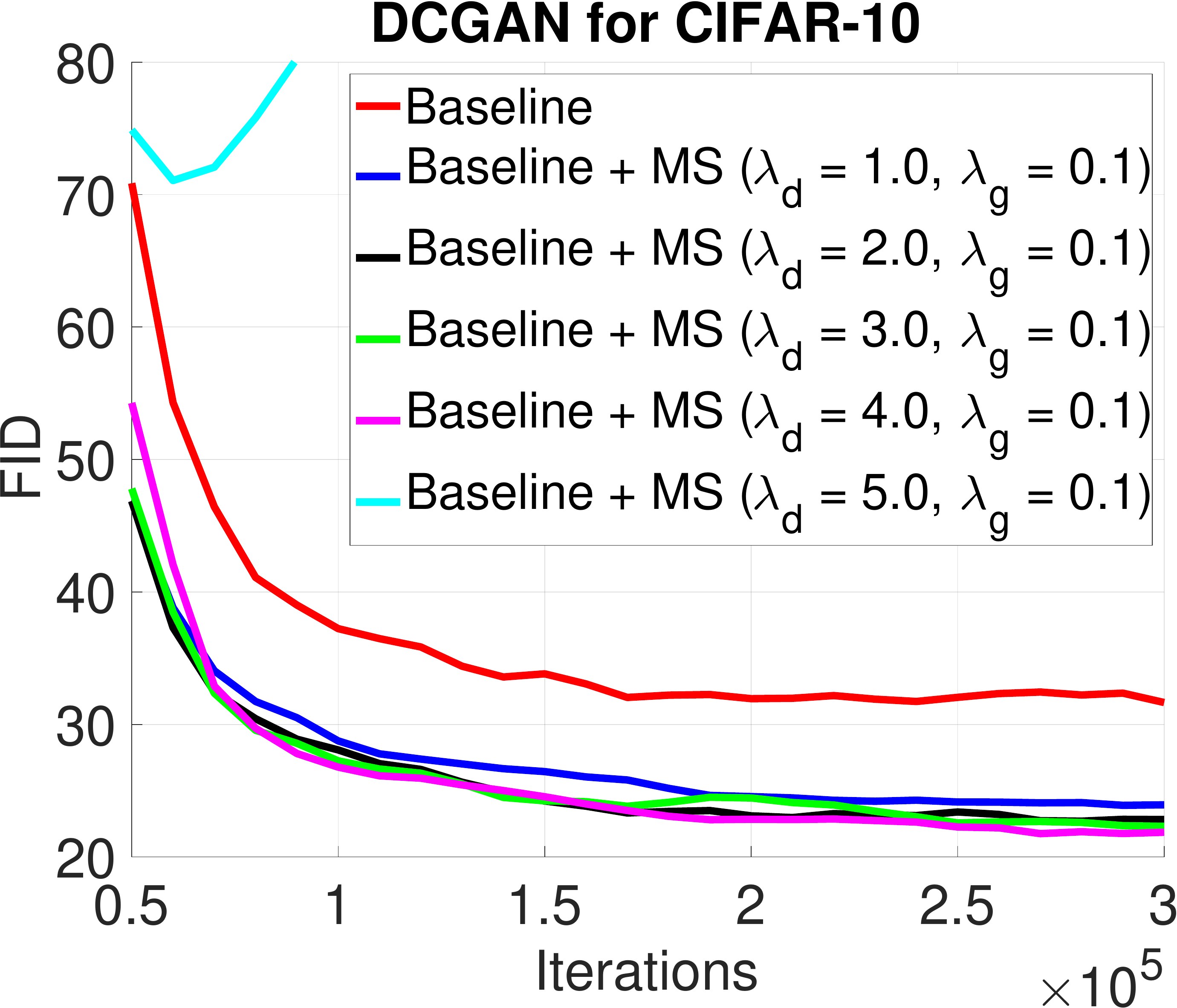}
  \caption{Our model (Dist-GAN + MS) with (a) with fine-tuning $\lambda_g$, fixed $\lambda_d = 1.0$. (b) fine-tuning $\lambda_d$, fixed $\lambda_g = 0.1$. The baseline is Dist-GAN model, and we use DCGAN architecture. (Best view in color)}
  \label{cifar_dcgan_finetuning_g_d}
\end{figure}

We first change the $\lambda_g$ according $\lambda_d = 1.0$ (Fig. \ref{cifar_dcgan_finetuning_g_d}a). With minimax game, the result suggests that $\lambda_g = 0.1$ is the best for DCGAN architecture. Then, we seek $\lambda_d$ with this $\lambda_g = 0.1$ as shown in Fig. \ref{cifar_dcgan_finetuning_g_d}b. Interestingly, now the best $\lambda_d = 4.0$, which is higher than $\lambda_d = 1.0$ of the original SS (the best with the original SS; Fig. \ref{ss_finetuning_dcgan}d). This suggests that using our proposed mini-max game based SS enable larger range of $\lambda_d$ with stable performance. 


\subsection{Ablation study of our proposed method with SN-GAN and ResNet architectures}
\label{appendix_state_of_the_art}

\begin{figure}
  \centering
  \includegraphics[width=3.3cm,keepaspectratio]{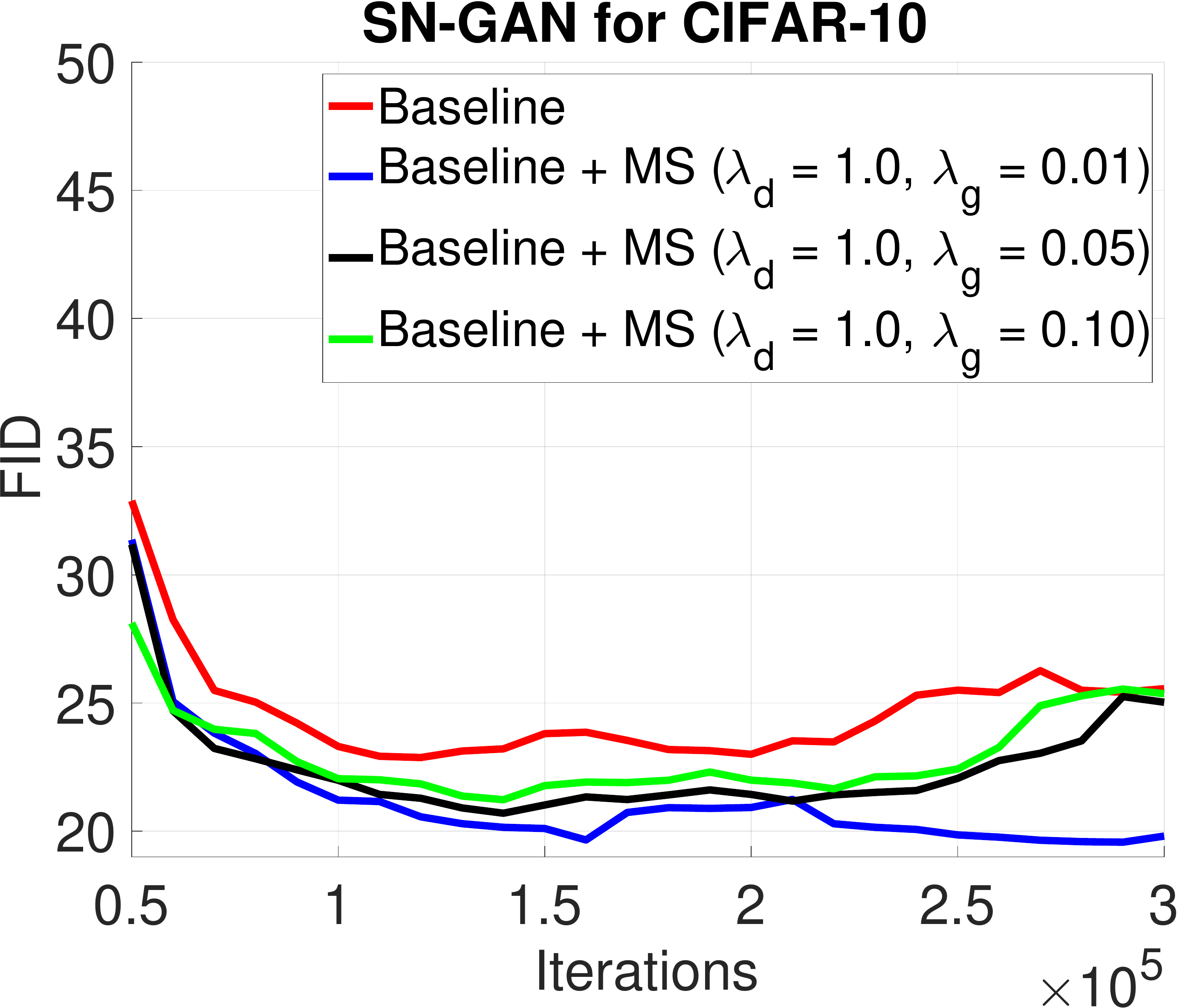}
  \includegraphics[width=3.3cm,keepaspectratio]{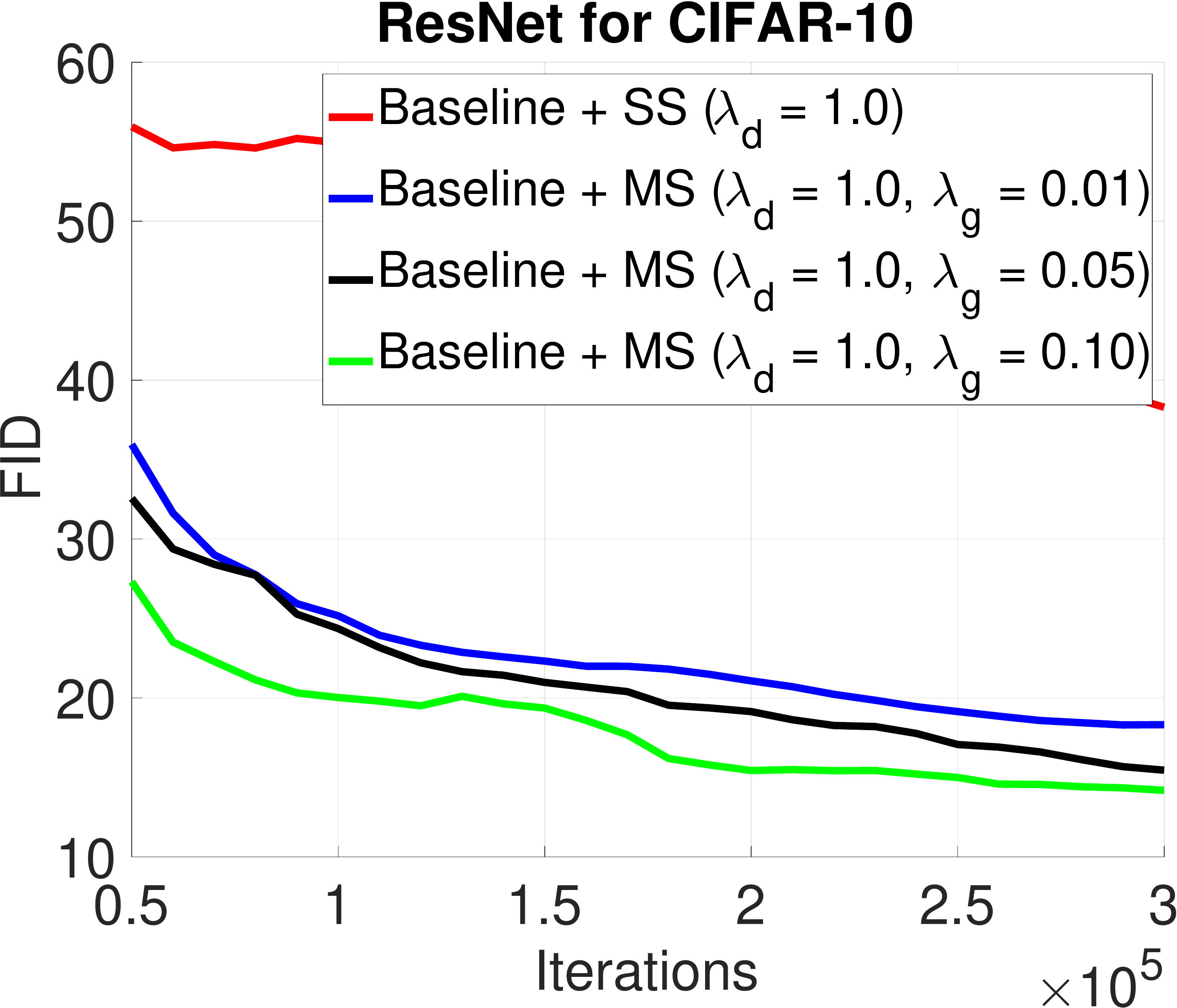}
  \includegraphics[width=3.3cm,keepaspectratio]{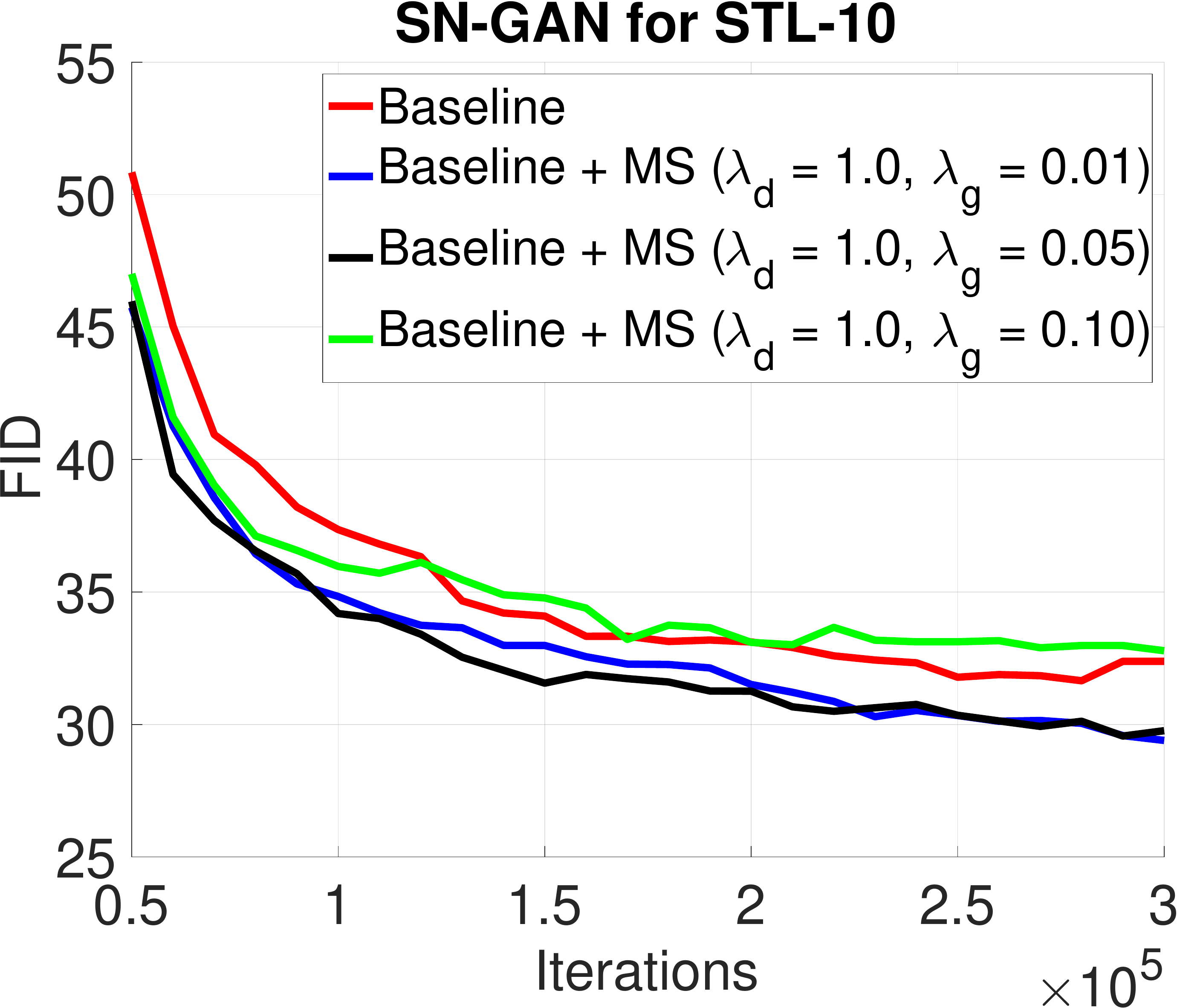}
  \includegraphics[width=3.3cm,keepaspectratio]{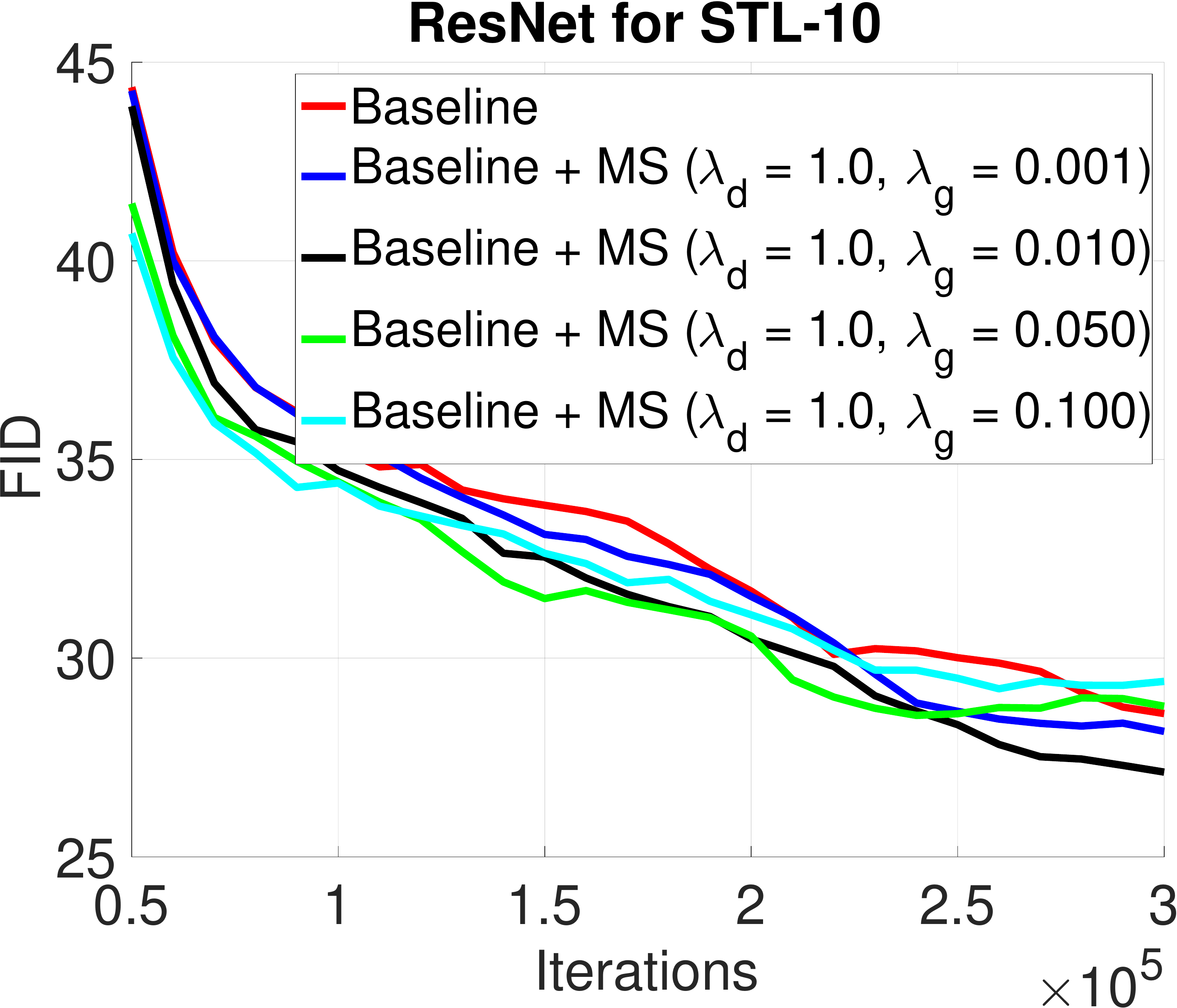}  
  \includegraphics[width=3.3cm,keepaspectratio]{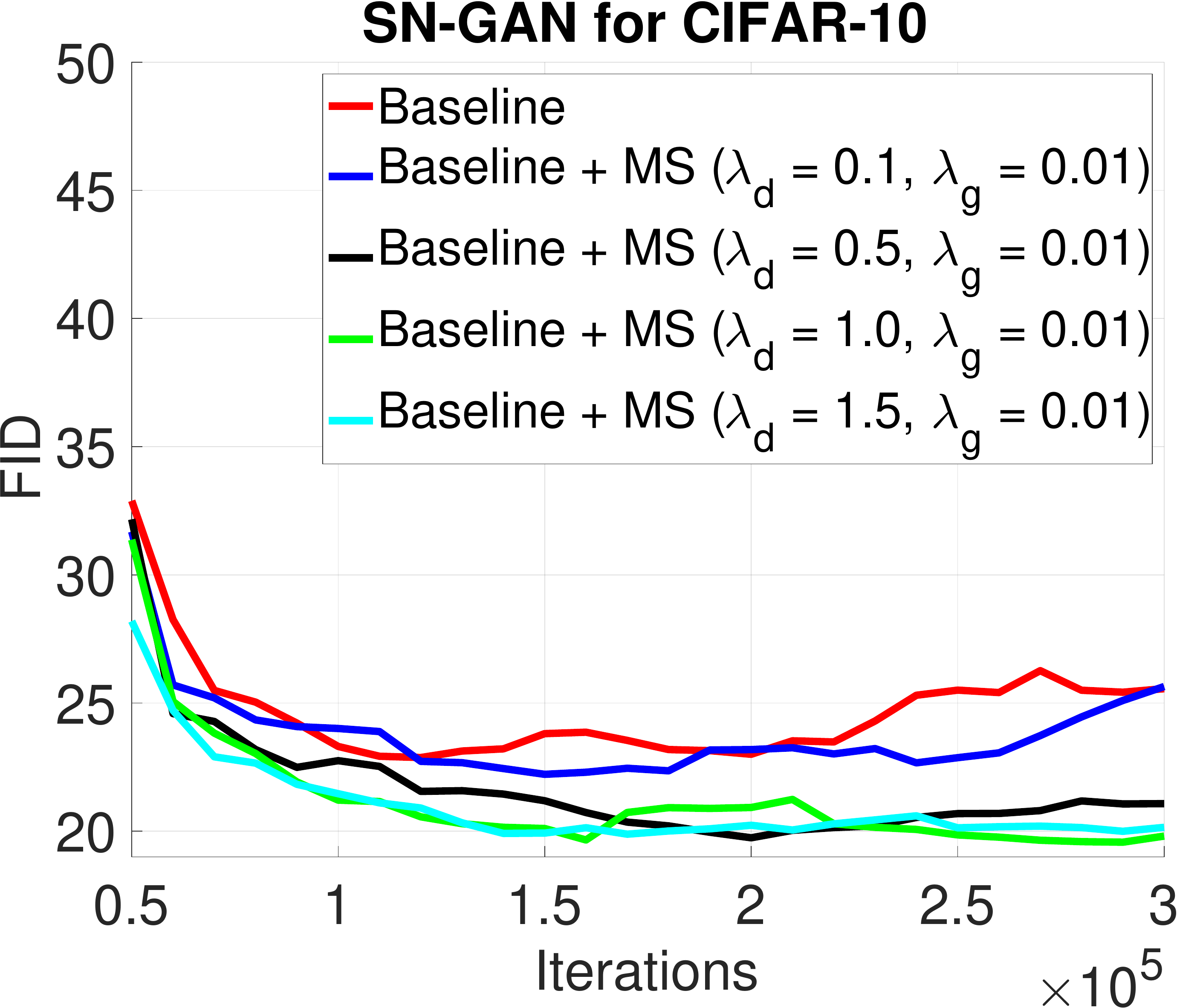}
  \includegraphics[width=3.3cm,keepaspectratio]{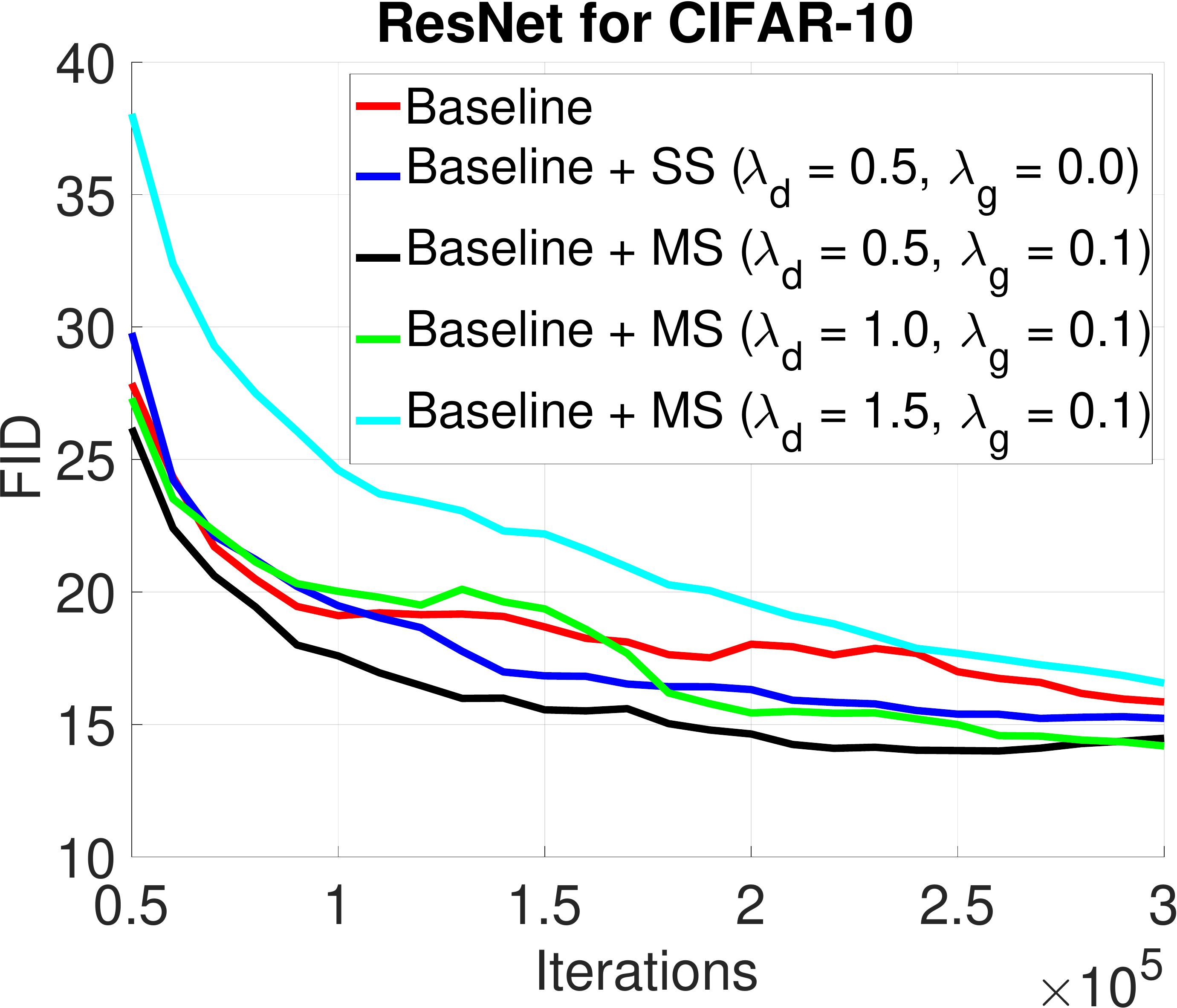}
  \includegraphics[width=3.3cm,keepaspectratio]{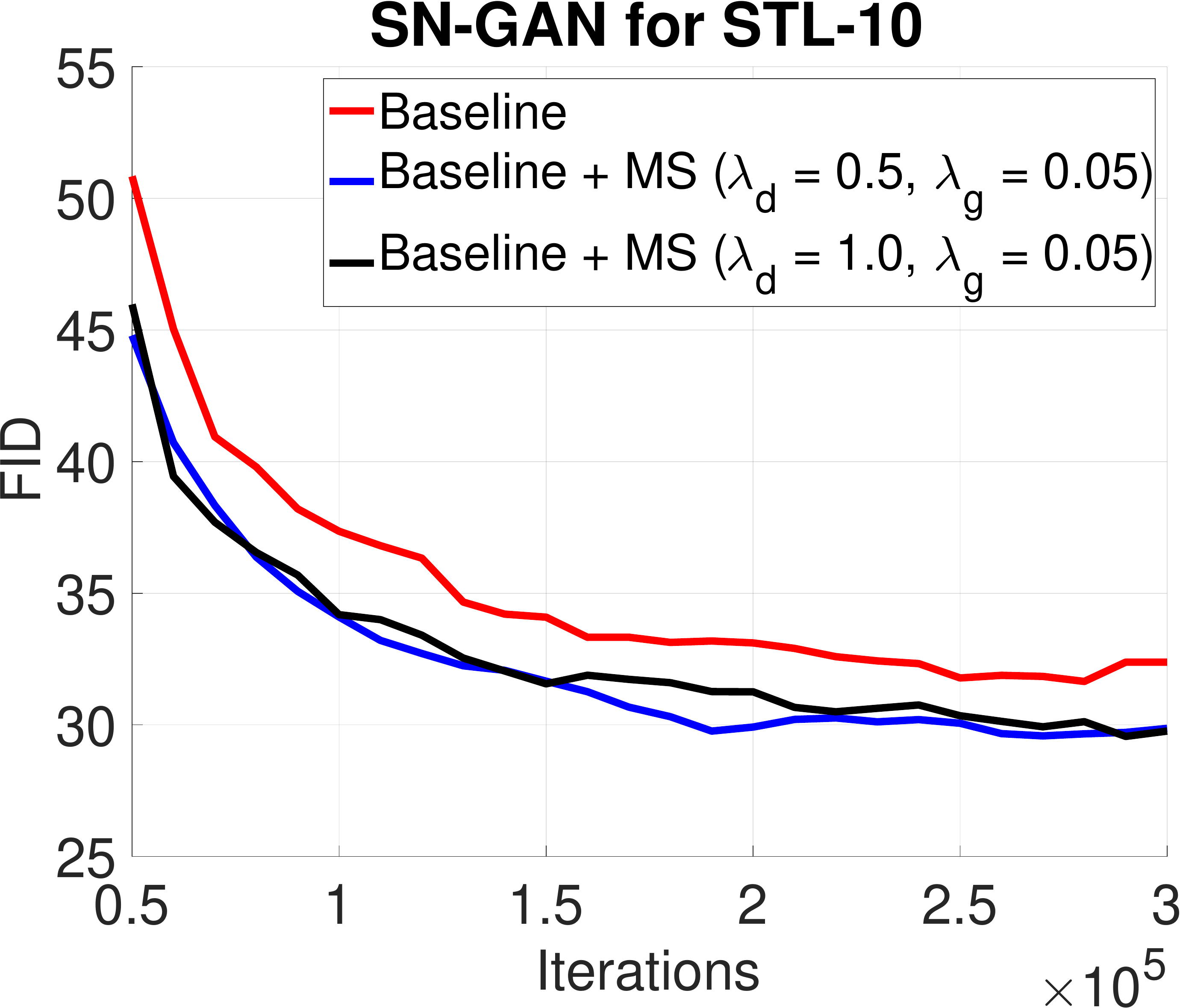}
  \includegraphics[width=3.3cm,keepaspectratio]{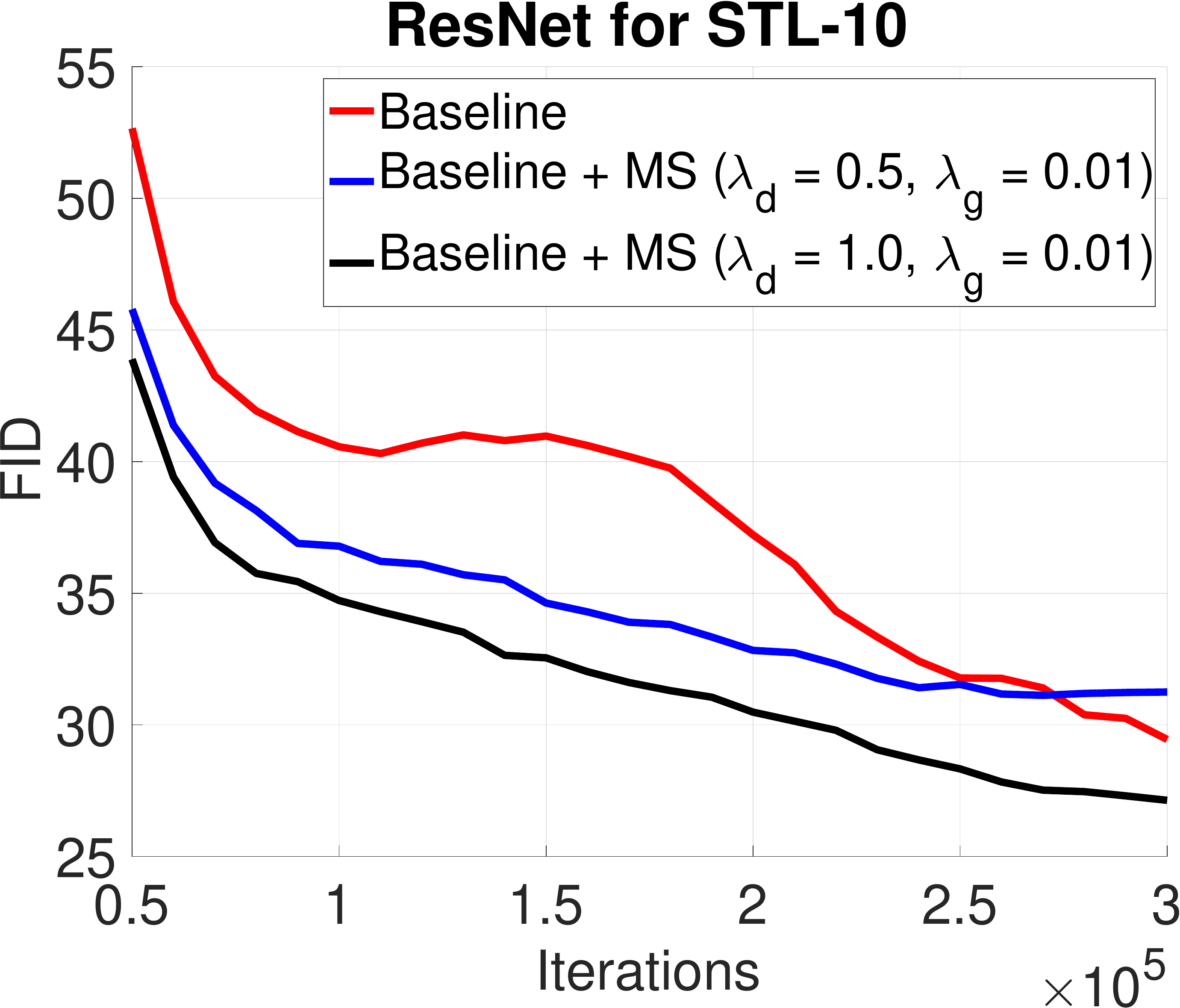}
  \caption{Understanding the effects of MS tasks (our proposed self-supervised tasks), by  fine-tuning $\lambda_g$ (first row) $\lambda_d$ (second row) for CIFAR-10 and STL-10 with other architectures. From left to right: SN-GAN for CIFAR-10, ResNet for CIFAR-10, SN-GAN for STL-10 and ResNet for STL-10. SN-GAN architecture referred as CNN architectures used in SN-GAN \cite{miyato-iclr-2018}.}
  \label{appendix_ss_d_g_finetuning_all} 
\end{figure}

The detail of the ablation study of $\lambda_d$ and $\lambda_g$ for our proposed SS tasks using SN-GAN and ResNet architectures are shown in Fig. \ref{appendix_ss_d_g_finetuning_all}.

\section{Appendix: Network architectures}

\label{network-architecture}

\subsection{DCGAN architecture}

Our DCGAN architecture, which is used for ablation studies on CIFAR-10, are presented in Table. \ref{dcgan}.

\begin{table}[ht!]
	\caption{\label{dcgan}Our DCGAN architecture is similar to \cite{radford-arxiv-2015} but the smaller number of feature maps (D = 64) to be more efficient for our ablation study on CIFAR-10. The Encoder is the mirror of the Generator. Slopes of lReLU functions are set to $0.2$. $\mathcal{U}(0, 1)$ is the uniform distribution.}
   	\centering
   	\scriptsize
    \begin{subtable}{.32\linewidth}
    	\centering
    	{\begin{tabular}{c}
			\toprule
			\midrule
		 	RGB image $x\in \bbR^{M\times M \times 3}$ \\
            \midrule
            5$\times$5, stride=2 conv. 1 $\times$ D ReLU\\
            \midrule
            5$\times$5, stride=2 conv. BN 2 $\times$ D ReLU\\        		 	
			\midrule
            5$\times$5, stride=2 conv. BN 4 $\times$ D ReLU\\            	
            \midrule
            5$\times$5, stride=2 conv. BN 8 $\times$ D ReLU\\    
            \midrule          
            dense $\rightarrow$ 128 \\
            \midrule
			\bottomrule
		\end{tabular}}
        \caption{\label{tab:enc_dcgan}Encoder, $M=32$ for CIFAR-10}
    \end{subtable}   	
    \begin{subtable}{.32\linewidth}
    	\centering
    	{\begin{tabular}{c}
			\toprule
			\midrule
		 	$z\in \bbR^{128} \sim \mathcal{U}(0, 1)$ \\	 	
           	\midrule
            dense $\rightarrow$ 2 $\times$ 2 $\times$ 8 $\times$ D  \\
            \midrule
            5$\times$5, stride=2 deconv. BN 4 $\times$ D ReLU\\
			\midrule
            5$\times$5, stride=2 deconv. BN 2 $\times$ D ReLU\\
            \midrule
            5$\times$5, stride=2 deconv. BN 1 $\times$ D ReLU\\
            \midrule
            5$\times$5, stride=2 deconv. 3 Sigmoid\\	
            \midrule
			\bottomrule
		\end{tabular}}
        \caption{\label{tab:gen_dcgan}Generator for CIFAR-10}
    \end{subtable}
    \begin{subtable}{.34\linewidth}
    	\centering
    	{\begin{tabular}{c}
			\toprule
			\midrule
			RGB image $x\in \bbR^{M\times M \times 3}$ \\
            \midrule
            5$\times$5, stride=2 conv. 1 $\times$ D lReLU\\
            \midrule
            5$\times$5, stride=2 conv. BN 2 $\times$ D lReLU\\        		 	
			\midrule
            5$\times$5, stride=2 conv. BN 4 $\times$ D lReLU\\            	
            \midrule
            5$\times$5, stride=2 conv. BN 8 $\times$ D lReLU\\    
            \midrule
            dense $\rightarrow$ 1, dense $\rightarrow$ 5 (two heads) \\
            \midrule
			\bottomrule
		\end{tabular}}
        \caption{\label{tab:dis_dcgan}Discriminator, $M=32$ for CIFAR-10. Two heads for the real/fake discriminator and multi-class classifier.}
    \end{subtable}
\end{table}

\subsection{SNGAN architecture}

Our SN-GAN architecture referred as CNN architectures of \cite{miyato-iclr-2018} for CIFAR-10 and STL-10 datasets are presented in Table. \ref{tab:sngan_models}.

\begin{table}[ht!]
	\caption{\label{tab:sngan_models} Encoder, generator, and discriminator of standard CNN architectures for CIFAR-10 and STL-10 used in our experiments. We use similar architectures as ones in \cite{miyato-iclr-2018}. The Encoder is the mirror of the Generator. Slopes of lReLU functions are set to $0.1$. $\mathcal{U}(0, 1)$ is the uniform distribution.}
   	\centering
   	\scriptsize
    \begin{subtable}{.33\linewidth}
    	\centering
    	{\begin{tabular}{c}
			\toprule
			\midrule
		 	RGB image $x\in \bbR^{M\times M \times 3}$ \\
            \midrule
            3$\times$3, stride=1 conv. 64\\	
            \midrule
            4$\times$4, stride=2 conv. BN 128 ReLU\\            		 	
			\midrule
            4$\times$4, stride=2 conv. BN 256 ReLU\\           	
            \midrule
            4$\times$4, stride=2 conv. BN 512 ReLU\\  
            \midrule          
            dense $\rightarrow$ 128 \\
            \midrule
			\bottomrule
		\end{tabular}}
        \caption{\label{tab:enc}Encoder, $M=32$ for CIFAR-10, and $M=48$ for STL-10}
    \end{subtable}   	
    \begin{subtable}{.33\linewidth}
    	\centering
    	{\begin{tabular}{c}
			\toprule
			\midrule
		 	$z\in \bbR^{128} \sim \mathcal{U}(0, 1)$ \\	 	
           	\midrule
            dense $\rightarrow$ $M_g$ $\times$ $M_g$ $\times$ 512 \\
            \midrule
            4$\times$4, stride=2 deconv. BN 256 ReLU\\
			\midrule
            4$\times$4, stride=2 deconv. BN 128 ReLU\\
            \midrule
            4$\times$4, stride=2 deconv. BN 64 ReLU\\
            \midrule
            3$\times$3, stride=1 conv. 3 Sigmoid\\	
            \midrule
			\bottomrule
		\end{tabular}}
        \caption{\label{tab:gen}Generator, $M_g=4$  for CIFAR-10, and $M_g=6$ for STL-10}
    \end{subtable}
    \begin{subtable}{.32\linewidth}
    	\centering
    	{\begin{tabular}{c}
			\toprule
			\midrule
		 	RGB image $x\in \bbR^{M\times M \times 3}$ \\
			\midrule
            3$\times$3, stride=1 conv 64 lReLU\\
            4$\times$4, stride=2 conv 64 lReLU\\
            \midrule
            3$\times$3, stride=1 conv 128 lReLU\\
            4$\times$4, stride=2 conv 128 lReLU\\
            \midrule
            3$\times$3, stride=1 conv 256 lReLU\\
            4$\times$4, stride=2 conv 256 lReLU\\
            \midrule
            3$\times$3, stride=1 conv. 512 lReLU\\
            \midrule
            dense $\rightarrow$ 1, dense $\rightarrow$ 5 (two heads)\\
			\bottomrule
		\end{tabular}}
        \caption{\label{tab:dis_deep}Discriminator, $M=32$ for CIFAR-10, and $M=48$ for STL-10. Two heads for the real/fake discriminator and multi-class classifier.}
    \end{subtable}
\end{table}

\subsection{ResNet architecture}

Our ResNet architectures for CIFAR-10 and STL-10 are presented in Table. \ref{tab:resnets_cifar10} and Table. \ref{tab:resnets_stl}.

\begin{figure}[ht!]
	\begin{tabular}{cc}
	     \scriptsize
        \begin{minipage}{1.\textwidth}
          \tblcaption{\label{tab:resnets_cifar10}ResNet architecture for CIFAR10 dataset. The Encoder is the mirror of the Generator. We use similar architectures and ResBlock to the ones used in \cite{miyato-iclr-2018}. $\mathcal{U}(0, 1)$ is the uniform distribution.}
          \centering
          \begin{subtable}{.33\textwidth}
                        \centering
                        {\begin{tabular}{c}
                            \toprule
                            \midrule
                            RGB image $x\in \bbR^{32\times 32 \times 3}$ \\
                            \midrule
                            3$\times$3 stride=1, conv. 256\\ 
                            \midrule
                            ResBlock down 256\\                
                            \midrule
                            ResBlock down 256\\
                            \midrule
                            ResBlock down 256\\                                                                   
                            \midrule
                            dense $\rightarrow$ 128 \\
                            \midrule
                            \bottomrule
                        \end{tabular}}
                        \caption{Encoder}
                    \end{subtable}
          \begin{subtable}{.33\textwidth}
              \centering
              {\begin{tabular}{c}
                  \toprule
                  \midrule
                  $z\in \bbR^{128} \sim \mathcal{U}(0, 1)$ \\
                  \midrule
                  dense, $4 \times 4 \times 256$ \\
                  \midrule
                  ResBlock up 256\\
                  \midrule
                  ResBlock up 256\\
                  \midrule
                  ResBlock up 256\\
                  \midrule
                  BN, ReLU, 3$\times$3 conv, 3 Sigmoid\\
                  \midrule
                  \bottomrule
              \end{tabular}}
              \caption{Generator}
          \end{subtable}
          \begin{subtable}{.32\textwidth}
              \centering
              {\begin{tabular}{c}
                  \toprule
                  \midrule
                  RGB image $x\in \bbR^{32\times 32 \times 3}$ \\
                  \midrule
                  ResBlock down 128\\
                  \midrule
                  ResBlock down 128\\
                  \midrule
                  ResBlock 128\\
                  \midrule
                  ResBlock 128\\
                  \midrule
                  ReLU\\
                  \midrule
                  Global sum pooling\\
                  \midrule
                  dense $\rightarrow$ 1, dense $\rightarrow$ 5 (two heads)\\
                  \midrule
                  \bottomrule
              \end{tabular}}
              \caption{Discriminator. Two heads for the real/fake discriminator and multi-class classifier.}
          \end{subtable}
        \end{minipage}
    \end{tabular}
\end{figure}

\begin{table}[ht!]
          \caption{\label{tab:resnets_stl}ResNet architecture for STL-10 dataset. The Encoder is the mirror of the Generator. We use similar architectures and ResBlock to the ones used in \cite{miyato-iclr-2018}. $\mathcal{U}(0, 1)$ is the uniform distribution.}
          \centering
          \scriptsize
          \begin{subtable}{.33\textwidth}
                        \centering
                        {\begin{tabular}{c}
                            \toprule
                            \midrule
                            RGB image $x\in \bbR^{48\times 48 \times 3}$\\
                            \midrule
                            3$\times$3 stride=1, conv. 64\\                            
                            \midrule
                            ResBlock down 128\\
                            \midrule 
                            ResBlock down 256\\
                            \midrule 
                            ResBlock down 512\\
                            \midrule                                 
                            dense $\rightarrow$ 128 \\
                            \midrule
                            \bottomrule
                        \end{tabular}}
                        \caption{Encoder}
                    \end{subtable}
          \begin{subtable}{.33\textwidth}
              \centering
              {\begin{tabular}{c}
                  \toprule
                  \midrule
                  $z\in \bbR^{128} \sim \mathcal{U}(0, 1)$ \\
                  \midrule
                  dense, $6 \times 6 \times 512$ \\
                  \midrule
                  ResBlock up 256\\
                  \midrule
                  ResBlock up 128\\
                  \midrule
                  ResBlock up 64\\
                  \midrule
                  BN, ReLU, 3$\times$3 conv, 3 Sigmoid\\
                  \midrule
                  \bottomrule
              \end{tabular}}
              \caption{Generator}
          \end{subtable}
          \begin{subtable}{.32\textwidth}
              \centering
              {\begin{tabular}{c}
                  \toprule
                  \midrule
                  RGB image $x\in \bbR^{48\times 48 \times 3}$ \\
                  \midrule
                  ResBlock down 64\\
                  \midrule
                  ResBlock down 128\\
                  \midrule
                  ResBlock down 256\\
                  \midrule
                  ResBlock down 512\\
                  \midrule
                  ResBlock 1024\\
                  \midrule
                  ReLU\\
                  \midrule
                  Global sum pooling\\
                  \midrule
                  dense $\rightarrow$ 1, dense $\rightarrow$ 5 (two heads)\\
                  \midrule
                  \bottomrule
              \end{tabular}}
              \caption{Discriminator. Two heads for the real/fake discriminator and multi-class classifier.}
          \end{subtable}
\end{table}

\end{document}